%% file: main.tex
\documentclass[lettersize, journal]{IEEEtran}

\usepackage{amsmath,amssymb,amsmath,amsfonts}
\usepackage{bm}
\usepackage{bbm}
\usepackage{graphicx}
\usepackage{cite}
\usepackage{epstopdf}
\usepackage{balance}
\usepackage{gensymb}
\usepackage{color}
\usepackage{lipsum}
\usepackage{float}
\usepackage{epstopdf}
\usepackage[mathcal]{eucal}
\usepackage{subfiles}
\usepackage[T1]{fontenc}
\usepackage{siunitx}
\usepackage{tabulary}

\usepackage{epsfig,graphics,graphicx,amssymb,amstext,amsmath,algorithm,algorithmic,wrapfig,multirow}
\usepackage{array}
\usepackage{adjustbox}
\usepackage[dvipsnames]{xcolor}
\usepackage{cite}
\usepackage{times}
\usepackage{placeins}
\usepackage{textcomp}
\usepackage{flushend}
\usepackage{xcolor}
\usepackage{tabularx}
\usepackage{bm}
\usepackage{enumerate}
\usepackage{tikz}
\usepackage{pgfplots}
\usepackage{epstopdf}
\usetikzlibrary{shapes,arrows}
\usepackage[utf8]{inputenc}
\usepackage{mathtools}
\usepackage[utf8]{inputenc}
\usepackage{xcolor}
\usepackage{tikz}
\usetikzlibrary{chains,arrows,calc,positioning}
\usepackage{todonotes}

\pgfplotsset{compat=1.16}
\definecolor{bittersweet}{rgb}{1.0, 0.44, 0.37}
\definecolor{glaucous}{rgb}{0.38, 0.51, 0.71}
\definecolor{gainsboro}{rgb}{0.86, 0.86, 0.86}
\definecolor{babyblueeyes}{rgb}{0.63, 0.79, 0.95}
\definecolor{silver}{rgb}{0.75, 0.75, 0.75}
\definecolor{neoncarrot}{rgb}{1.0, 0.64, 0.26}

\usepackage{soul}

\usepackage{breqn}
\usepackage{booktabs}
\usepackage{multirow}

\usepackage{enumitem}
\usepackage{tabulary}

\usepackage[normalem]{ulem}

\input{acronyms.tex}

\makeatletter

 \let\oldforeign@language\foreign@language
 \DeclareRobustCommand{\foreign@language}[1]{%
   \lowercase{\oldforeign@language{#1}}}

\usepackage[caption=false,font=normalsize,labelfont=sf,textfont=sf]{subfig}

\usepackage{dblfloatfix}

\input{commands.tex}

\makeatother

\IEEEoverridecommandlockouts

\begin{document}
%
\title{Millimeter Wave 
Wireless Assisted Robot Navigation with
Link State Classification}
\author{
%
\IEEEauthorblockN{Mingsheng Yin$^{a}$, Akshaj Kumar Veldanda$^{a}$, Amee Trivedi$^{b}$, Jeff Zhang$^{c}$, Kai Pfeiffer$^{a}$, Yaqi Hu$^{a}$,\\ Siddharth Garg$^{a}$, Elza Erkip$^{a}$, Ludovic Righetti$^{a}$, Sundeep Rangan$^{a}$} \\
\IEEEauthorblockA{$^{a}$NYU Tandon School of Engineering, Brooklyn, NY, USA\\
$^{b}$University of British Columbia, Vancouver, BC, Canada\\
$^{c}$Harvard University, Cambridge, MA, USA\\}

\thanks{The authors were supported 
by NSF grants 
1952180, 1925079, 1564142, 1547332, the SRC, OPPO,
and the industrial affiliates of NYU WIRELESS.  The work was also supported by RemCom that provided the Wireless Insite  
software.}
}

\maketitle
\begin{abstract}
\input{abstract.tex}
\end{abstract}


\IEEEpeerreviewmaketitle

\input{intro.tex}

\input{problem}
\input{algo}

\input{raytracing}
\input{sim}

\input{navigation}
\input{conclusion}


\bibliographystyle{IEEEtran}
\bibliography{bibl}

\end{document}

%% file: acronyms.tex
\usepackage[acronyms,nonumberlist,nopostdot,nomain,nogroupskip]{glossaries}
\newacronym{quic}{QUIC}{Quick UDP Internet Connections}
\newacronym{3gpp}{3GPP}{3rd Generation Partnership Project}
\newacronym{adc}{ADC}{Analog to Digital Converter}
\newacronym{5g}{5G}{5th generation}
\newacronym{aimd}{AIMD}{Additive Increase Multiplicative Decrease}
\newacronym{am}{AM}{Acknowledged Mode}
\newacronym{amc}{AMC}{Adaptive Modulation and Coding}
\newacronym{aqm}{AQM}{Active Queue Management}
\newacronym{awgn}{AGWN}{Additive White Gaussian Noise}
\newacronym{afd}{AFD}{Austin Fire Department}
\newacronym{balia}{BALIA}{Balanced Link Adaptation}
\newacronym{bdp}{BDP}{Bandwidth-Delay Product}
\newacronym{bf}{BF}{Beamforming}
\newacronym{cc}{CC}{Congestion Control}
\newacronym{cdf}{CDF}{Cumulative Distribution Function}
\newacronym{cn}{CN}{Core Network}
\newacronym{cqi}{CQI}{Channel Quality Information}
\newacronym{cp}{CP}{Control Plane}
\newacronym{csirs}{CSI-RS}{Channel State Information - Reference Signal}
\newacronym{dc}{DC}{Dual Connectivity}
\newacronym{dce}{DCE}{Direct Code Execution}
\newacronym{dci}{DCI}{Downlink Control Information}
\newacronym{dl}{DL}{Downlink}
\newacronym{dmr}{DMR}{Deadline Miss Ratio}
\newacronym{dmrs}{DMRS}{DeModulation Reference Signal}
\newacronym{e2e}{E2E}{End-to-End}
\newacronym{ecn}{ECN}{Explicit Congestion Notification}
\newacronym{edf}{EDF}{Earliest Deadline First}
\newacronym{enb}{eNB}{evolved Node Base}
\newacronym{epc}{EPC}{Evolved Packet Core}
\newacronym{es}{ES}{Edge Server}
\newacronym{fdma}{FDMA}{Frequency Division Multiple Access}
\newacronym{fdd}{FDD}{Frequency Division Duplexing}
\newacronym[firstplural=Radio Access Technologies (RATs)]{rat}{RAT}{Radio Access Technology}
\newacronym{fs}{FS}{Fast Switching}
\newacronym{ftp}{FTP}{File Transfer Protocol}
\newacronym{gnb}{gNB}{Next Generation Node Base}
\newacronym{harq}{HARQ}{Hybrid Automatic Repeat reQuest}
\newacronym{hetnet}{HetNet}{Heterogeneous Network}
\newacronym{hh}{HH}{Hard Handover}
\newacronym{hol}{HOL}{Head-of-Line}
\newacronym{ia}{IA}{Initial Access}
\newacronym{imt}{IMT}{International Mobile Telecommunication}
\newacronym{iot}{IoT}{Internet of Things}
\newacronym{los}{LOS}{Line of Sight}
\newacronym{lte}{LTE}{Long Term Evolution}
\newacronym{m2m}{M2M}{Machine to Machine}
\newacronym{mac}{MAC}{Medium Access Control}
\newacronym{mc}{MC}{Multi-Connectivity}
\newacronym{mcs}{MCS}{Modulation and Coding Scheme}
\newacronym{mec}{MEC}{Mobile Edge Cloud}
\newacronym{mi}{MI}{Mutual Information}
\newacronym{mimo}{MIMO}{Multiple Input, Multiple Output}
\newacronym{mmwave}{mmWave}{millimeter wave}
\newacronym{mr}{MR}{Maximum Rate}
\newacronym{mss}{MSS}{Maximum Segment Size}
\newacronym{mtd}{MTD}{Machine-Type Device}
\newacronym{mtu}{MTU}{Maximum Transmission Unit}
\newacronym{nfv}{NFV}{Network Function Virtualization}
\newacronym{nlos}{NLOS}{Non Line of Sight}
\newacronym{nr}{NR}{New Radio}
\newacronym{ofdm}{OFDM}{Orthogonal Frequency Division Multiplexing}
\newacronym{pdcch}{PDCCH}{Physical Downlonk Control Channel}
\newacronym{pdcp}{PDCP}{Packet Data Convergence Protocol}
\newacronym{pdsch}{PDSCH}{Physical Downlink Shared Channel}
\newacronym{pdu}{PDU}{Packet Data Unit}
\newacronym{pf}{PF}{Proportional Fair}
\newacronym{pgw}{PGW}{Packet Gateway}
\newacronym{phy}{PHY}{Physical}
\newacronym{pbch}{PBCH}{Physical Broadcast Channel}
\newacronym[plural=\gls{mme}s,firstplural=Mobility Management Entities (MMEs)]{mme}{MME}{Mobility Management Entity}
\newacronym{prb}{PRB}{Physical Resource Block}
\newacronym{pss}{PSS}{Primary Synchronization Signal}
\newacronym{pucch}{PUCCH}{Physical Uplink Control Channel}
\newacronym{pusch}{PUSCH}{Physical Uplink Shared Channel}
\newacronym{rach}{RACH}{Random Access Channel}
\newacronym{ran}{RAN}{Radio Access Network}
\newacronym{red}{RED}{Robotics Emergency Deployment}
\newacronym{rf}{RF}{Radio Frequency}
\newacronym{rlc}{RLC}{Radio Link Control}
\newacronym{rlf}{RLF}{Radio Link Failure}
\newacronym{rrc}{RRC}{Radio Resource Control}
\newacronym{rrm}{RRM}{Radio Resource Management}
\newacronym{rr}{RR}{Round Robin}
\newacronym{rs}{RS}{Remote Server}
\newacronym{rsrp}{RSRP}{Reference Signal Received Power}
\newacronym{rss}{RSS}{Received Signal Strength}
\newacronym{rtt}{RTT}{Round Trip Time}
\newacronym{rw}{RW}{Receive Window}
\newacronym{rx}{RX}{Receiver}
\newacronym{sa}{SA}{standalone}
\newacronym{sack}{SACK}{Selective Acknowledgment}
\newacronym{sap}{SAP}{Service Access Point}
\newacronym{sch}{SCH}{Secondary Cell Handover}
\newacronym{scoot}{SCOOT}{Split Cycle Offset Optimization Technique}
\newacronym{sdma}{SDMA}{Spatial Division Multiple Access}
\newacronym{sinr}{SINR}{Signal to Interference plus Noise Ratio}
\newacronym{sm}{SM}{Saturation Mode}
\newacronym{snr}{SNR}{Signal to Noise Ratio}
\newacronym{son}{SON}{Self-Organizing Network}
\newacronym{ss}{SS}{Synchronization Signal}
\newacronym{srs}{SRS}{Sounding Reference Signal}
\newacronym{sss}{SSS}{Secondary Synchronization Signal}
\newacronym{tb}{TB}{Transport Block}
\newacronym{tcp}{TCP}{Transmission Control Protocol}
\newacronym{tdd}{TDD}{Time Division Duplexing}
\newacronym{tdma}{TDMA}{Time Division Multiple Access}
\newacronym{tfl}{TfL}{Transport for London}
\newacronym{tm}{TM}{Transparent Mode}
\newacronym{trp}{TRP}{Transmitter Receiver Pair}
\newacronym{tti}{TTI}{Transmission Time Interval}
\newacronym{ttt}{TTT}{Time-to-Trigger}
\newacronym{tx}{TX}{Transmitter}
\newacronym{ue}{UE}{User Equipment}
\newacronym{ul}{UL}{Uplink}
\newacronym{uml}{UML}{Unified Modeling Language}
\newacronym{um}{UM}{Unacknowledged Mode}
\newacronym{utc}{UTC}{Urban Traffic Control}
\newacronym{vm}{VM}{Virtual Machine}
\newacronym{rsrq}{RSRQ}{Reference Signal Received Quality}
\newacronym{rssi}{RSSI}{Received Signal Strength Indicator}
\newacronym{crs}{CRS}{Cell Reference Signal}
\newacronym{comp}{CoMP}{Coordinated Multi-Point}
\newacronym{cran}{C-RAN}{Cloud \acrlong{ran}}
\newacronym{ca}{CA}{Carrier Aggregation}
\newacronym{cco}{CC}{Carrier Component}
\newacronym{nsa}{NSA}{Non Stand Alone}
\newacronym{embb}{eMBB}{Enhanced Mobility Broadband}
\newacronym{bsr}{BSR}{Buffer Status Report}
\newacronym{srb}{SRB}{Service Radio Bearer}
\newacronym{scm}{SCM}{Spatial Channel Model}
\newacronym{sctp}{SCTP}{Stream Control Transmission Protocol}
\newacronym{mptcp}{MPTCP}{Multi-path TCP}
\newacronym{ietf}{IETF}{Internet Engineering Task Force}
\newacronym{os}{OS}{Operating System}
\newacronym{tls}{TLS}{Transport Layer Security}
\newacronym{rfc}{RFC}{Request for Comments}
\newacronym{http}{HTTP}{HyperText Transfer Protocol}
\newacronym{nat}{NAT}{Network Address Translation}
\newacronym{api}{API}{Application Programming Interface}
\newacronym{rto}{RTO}{Retransmission Timeout}
\newacronym{psc}{PSC}{Public Safety Communication}
\newacronym{rpgm}{RPGM}{Reference Point Group Mobility}
\newacronym{ic}{IC}{Incident Command}
\newacronym{rsu}{RSU}{Road Side Unit}
\newacronym{uav}{UAV}{unmanned aerial vehicle}
\newacronym{usv}{USV}{Unmanned Surface Vehicle}
\newacronym{uas}{UAS}{Unmanned Aerial System}
\newacronym{iab}{IAB}{Integrated Access and Backhaul}
\newacronym{qoe}{QoE}{Quality of Experience}
\newacronym{ssim}{SSIM}{Structural Similarity Index}
\newacronym{psnr}{PSNR}{Peak Signal to Noise Ratio}
\newacronym{bs}{BS}{Base Station}
\newacronym{mu}{MU}{Multiple User}
\newacronym{ag}{AG}{Air-to-Ground}
\newacronym{af}{AF}{Array Factor}
\newacronym{ula}{ULA}{Uniform Linear Array}
\newacronym{upa}{UPA}{Uniform Planar Array}
\newacronym{lcs}{LCS}{Local Coordinate System}
\newacronym{psd}{PSD}{Power Spectral Density}
\newacronym{vq}{VQ}{vector quantization}
\newacronym{a2g}{A2G}{air-to-ground}
\newacronym{em}{EM}{electromagnetic}
\newacronym{vae}{VAE}{variational autoencoder}

%% file: commands.tex
\def\nba{{\mathbf{a}}}
\def\nbb{{\mathbf{b}}}
\def\nbc{{\mathbf{c}}}

\def\nbr{{\mathbf{r}}}

\def\nbu{{\mathbf{u}}}
\def\nbv{{\mathbf{v}}}
\def\nbw{{\mathbf{w}}}

\def\nb0{{\mathbf{0}}}
\def\nb1{{\mathbf{1}}}

\def\argmax{\operatorname{arg~max}}
\def\Omegabar{\overline{\Omega}}
\def\taubar{\overline{\tau}}
\def\Omegahat{\widehat{\Omega}}
\def\tauhat{\widehat{\tau}}

%% file: abstract.tex
The millimeter wave (mmWave) bands have attracted considerable 
attention for high precision localization applications due to the ability to capture
high angular and temporal resolution
measurements.
This paper explores mmWave-based positioning for a target localization problem
where a fixed target broadcasts mmWave signals and a mobile robotic agent attempts 
to capture the signals to locate and navigate to the target.  A three-stage procedure
is proposed:  First, the mobile agent 
uses tensor decomposition methods to detect the multipath channel components and estimate their parameters.  
Second, 
a machine-learning trained classifier is then used to predict the \emph{link state}, meaning if the strongest
path is line-of-sight (LOS) or non-LOS (NLOS).
For the NLOS case, the link state predictor
also determines if the strongest path arrived 
via one or more reflections.  Third, based on the link state, the agent either follows the
estimated angles or uses computer vision or other sensor
to explore and map the environment.  The method is demonstrated on 
a large dataset of indoor environments supplemented with ray tracing 
to simulate the wireless propagation.  
The path estimation and link state classification are also integrated into a state-of-the-art neural 
simultaneous localization and mapping (SLAM) module to augment camera and LIDAR-based
navigation.
It is shown that the link state classifier can successfully
generalize to completely new environments outside the training set.  In addition,
the neural-SLAM module with the wireless
path estimation and link state classifier provides
rapid navigation to the target, close to 
a baseline that knows the target location.


%% file: intro.tex
\section{Introduction}
\label{sec:intro}

\IEEEPARstart{T}{he} millimeter wave (mmWave) bands have  
tremendous potential for new localization technologies ~\cite{guerra2015position, shahmansoori20155g, guidi2014millimeter} and have emerged as a central
component of the positioning methods in the
3GPP Fifth Generation New Radio (5G NR) standard 
\cite{3GPP38885,3GPP38857,3GPP21916,keating2019overview,dwivedi2021positioning}.
The wide bandwidths in the mmWave frequencies 
coupled with arrays with large number of elements
enable detecting paths with very high delay and angular resolution.

\begin{figure}
    \centering
    \includegraphics[width=6cm]{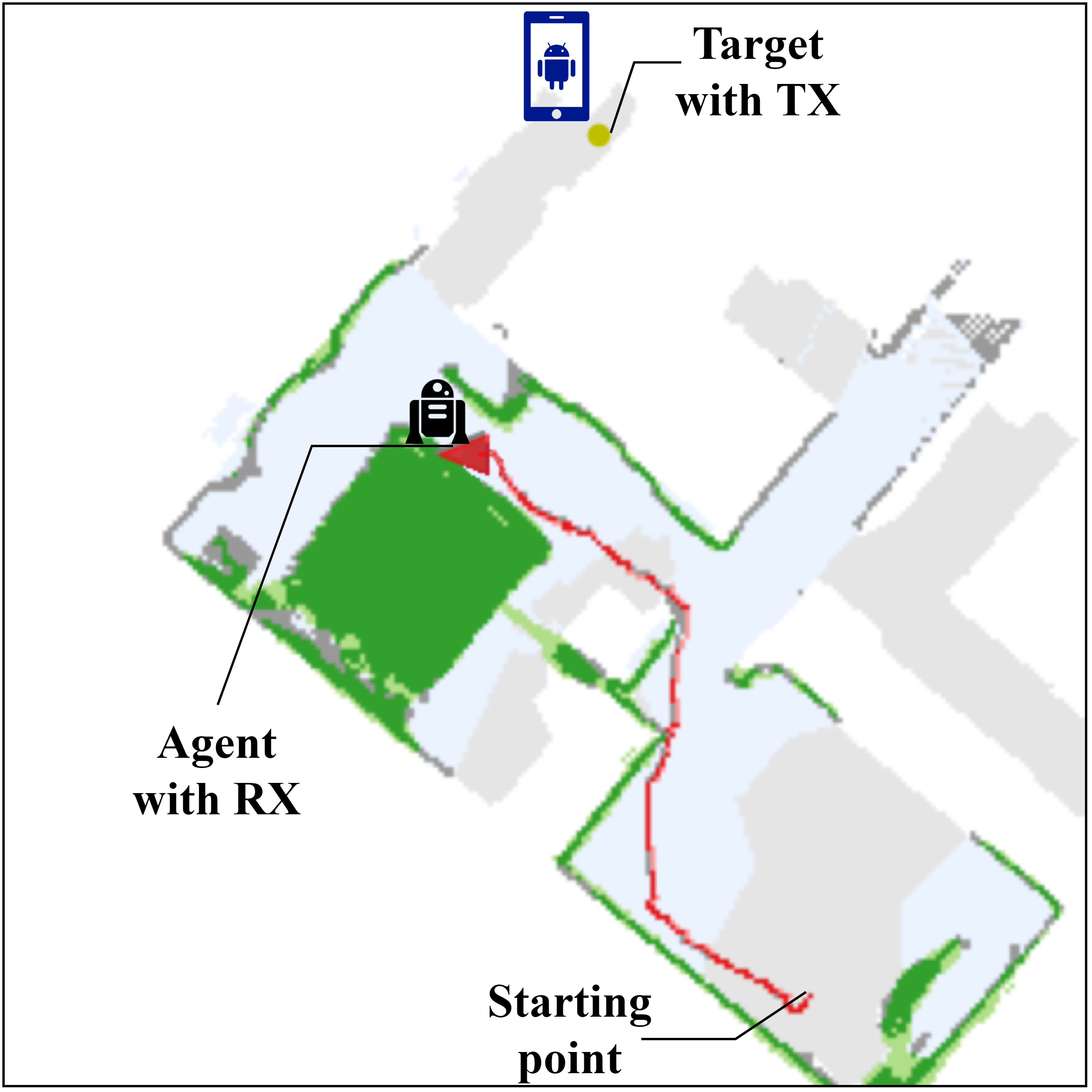}
    \caption{\textbf{Target localization and navigation}:
    A target with a wireless transponder and 
    a robotic agent
    must locate and navigate to the target
    using received wireless signals.  The path
    and map shown in the figure are example
    outputs of the Active Neural-SLAM module \cite{chaplot2020learning} augmented with
    the proposed mmWave wireless 
    path estimation and link state classification
    algorithm.}
    \label{fig:problem}
\end{figure}

This work considers mmWave-based positioning
in the context of a robotic target
localization application  motivated by search and rescue applications (illustrated in Fig.~\ref{fig:problem}):
In this problem,
a fixed target is located at some unknown position
and a mobile robotic 
agent must locate and navigate to the
target in the shortest possible time.  We assume the environment and obstacles are not known,
so the agent also needs to map and localize itself within the environment as 
part of the target discovery. 
To assist in the target localization, the target is equipped
with a mmWave transmitter that can periodically broadcast
known positioning signals.  The agent is equipped with a 
mmWave receiver that can attempt to locate the target from 
measurements of the positioning signals.
Specifically, the agent can attempt to estimate
the angles of arrivals of the signals to determine the direction in which to navigate so as to reach the target.
Note that, while ultra-wide
band (UWB) localization systems in traditional frequencies
such as \cite{zwirello2012uwb,zhang2008high} can achieve
high temporal resolution, mmWave is uniquely able to obtain
high angular resolution from a single receiver.  As we will see,
this angular information is critical for the target localization.

A key challenge in using angle of arrival estimation in such navigation applications 
is that
the strongest signal 
may not arrive from a line-of-sight
(LOS) path. 
In the mmWave bands in particular,
signals are highly susceptible to blocking
\cite{slezak2018empirical,Rappaport2014-mmwbook}. As such, strongest paths often arrive via 
non-LOS (NLOS) routes that might involve one or more reflections or diffractions. Following paths that involve many reflections/diffractions might not lead to the shortest route to the target.
To use the angular information, it is thus vital to first perform some form of \emph{link state classification} to estimate which measurements are in a LOS or NLOS state
\cite{guvenc2007nlos, venkatesh2007non,khodjaev2010survey}.
While there is a large body of work in mmWave positioning
\cite{wang2018pursuance,lu2020positioning,kim20185g},
link state classification in these frequencies is less
understood.  Positioning measurements in  
the mmWave range have several unique features including the need to support
directional antenna elements, multiple arrays, 
and beam sweeping which provide more complex set of inputs for link classification.  
The broad goal of this work is to understand
how to develop good path estimation and link state
classification algorithms that can be applied in the
context of target localization and navigation for realistic mmWave front-ends
and signaling. 

For this link state classification and target localization problem, we propose a three-step
procedure:  First, at each agent location, 
we use a modification of a
low-rank tensor decomposition algorithm 
\cite{zhou2017low,wen2018tensor} to detect
signal paths and estimate their path parameters
such as the angles of arrival.
Second, we train a neural network 
to determine if the strongest path is 
LOS or NLOS from the
path parameters. For NLOS paths, the network also 
determines if the path has arises from a 
single reflection or higher order reflections.
Importantly, the neural network is trained
on data distinct from the test environment,
so no prior knowledge or calibration in the 
environment is required.

Finally, for the target navigation, we employ the following simple algorithm:
When a link is classified as a LOS, strong single
reflection, or strong single diffraction, the agent simply follows the direction
of the strongest detected path. 
Otherwise, in absence of strong LOS or first-order reflection,
the
agent explores the environment using conventional
SLAM methods, potentially using other sensor modalities
such as camera or LiDAR.

The proposed method and analysis
has a number of features
that improve upon the state of the art:
\begin{itemize}
    \item \emph{Detailed antenna and multiple array modeling}:
    Practical mmWave devices at the terminal (UE)
    and base station (gNB) often use 
    multiple arrays oriented in different directions
    to obtain 360 degree coverage
    \cite{raghavan2019statistical,xia2020multi}.
    This work models these 
    multiple array structures and also includes
    detailed models of the 
    antenna element directivity in each array.  
    In addition, we
    do not consider any local oscillator (LO)
    synchronization across different arrays.
    
    \item \emph{Beam sweeping double directional estimation:}
    Many prior mmWave localization studies have either
    abstracted the directional estimation 
    \cite{kim20185g}, considered single-sided directional
    estimates \cite{wang2018pursuance,lu2020positioning}, 
    or considered 
    double directional estimates using MIMO signaling \cite{guerra2018single,huang2020machine}.
    In this work, we modify the low-rank tensor decomposition algorithms in 
    \cite{zhou2017low,wen2018tensor} to account
    for both sweeping of the TX beams and use of 
    multiple arrays at the TX and RX.
    Beam sweeping 
    at the transmitter is critical to model
    for most cellular mmWave systems \cite{heath2016overview}.
    
    \item \emph{Novel neural network feature input:} 
    Prior works such as \cite{huang2020machine} have used
    machine learning-based link state classification using
    aggregate features of the paths such as the spread of the received power, delays or angles of
    the paths.
    In this work, we propose a neural network that takes
    the raw features of each detected path.  This larger 
    feature space is enabled by training the network
    on extensive data from ray tracing.

    \item \emph{Multi-class link classification:}
    Instead of simply classifying the link as LOS or NLOS,
    we differentiate between four states:  LOS, 
    NLOS from a single interaction, higher-order NLOS
    and outage.
    We show that for target localization application,
    both LOS and first-order NLOS paths have angles
    of arrival that
    strongly correlate with good navigation directions.

    \item \emph{Validation in realistic, complex
    environments:}   The proposed approach is trained and tested on a large set 
    of detailed indoor environments from the Gibson data set \cite{xia2018gibson},
    that is widely-used in robotic navigation~\cite{savva2019habitat}.
    Each 3D model in the data set is imported to a state-of-the-art ray tracer
     \cite{Remcom}, where we predict the wireless paths 
    at \SI{28}{GHz}.   The wireless propagation data is then combined
    with realistic models of the arrays including the array sectorization 
    and orientation and estimation algorithms.

    \item \emph{Intra-site generalization ability:}  We demonstrate
    that models trained on one set of environments
    generalize to completely new environments.  Indeed,
    we show that, on new environments not in the training set,
    our link classification has a 88\% accuracy.  
    
    \item \emph{Target localization demonstration:}  The link state classification
    and angular estimation is combined with a state-of-the-art neural SLAM module
    \cite{chaplot2020learning} tested in AI Habitat robotic simulation environment \cite{savva2019habitat} with the Gibson dataset above \cite{xia2018gibson}.   
    In addition to the detailed 3D models, the dataset includes LIDAR and camera data which are the typical inputs of the SLAM methods.
    We simulate the simple navigation policy
    that follows the estimated path when the detected link state is LOS or first-order
    NLOS; otherwise, the agent uses the exploration mode in the neural SLAM.
    We believe this is the first full end-to-end detailed simulation of mmWave
    wireless localization combined with state-of-the-art neural SLAM.
    In addition, we show that the proposed multi-class link state classifier significantly
    outperforms a policy that only uses the LOS classification.
    
    \item \emph{Public dataset:}  All the code
    and wireless data is made public, providing the 
    first complete 5G wireless localization dataset
    combined with camera data and robotic simulation
    environment \cite{mmwRobotNav}.

\end{itemize}

%% file: problem.tex
\section{Path Estimation and Link State Classification Problem}
\label{sec:direst}

\subsection{Overview}
Before addressing the navigation problem specifically, in this section, 
we introduce a general problem 
of estimating multi-path components and
the link state from mmWave measurements.
To this end, 
consider a channel between 
a single TX and RX location.
We assume a ray model where the channel is
characterized by $L$ discrete paths
\cite{heath2018foundations}.
We will only consider paths with gains above
some threshold.  As we will see below,
the threshold is selected such that the path's
parameters can be estimated accurately.

\begin{figure}
  \centering
  \includegraphics[width=0.5\linewidth]{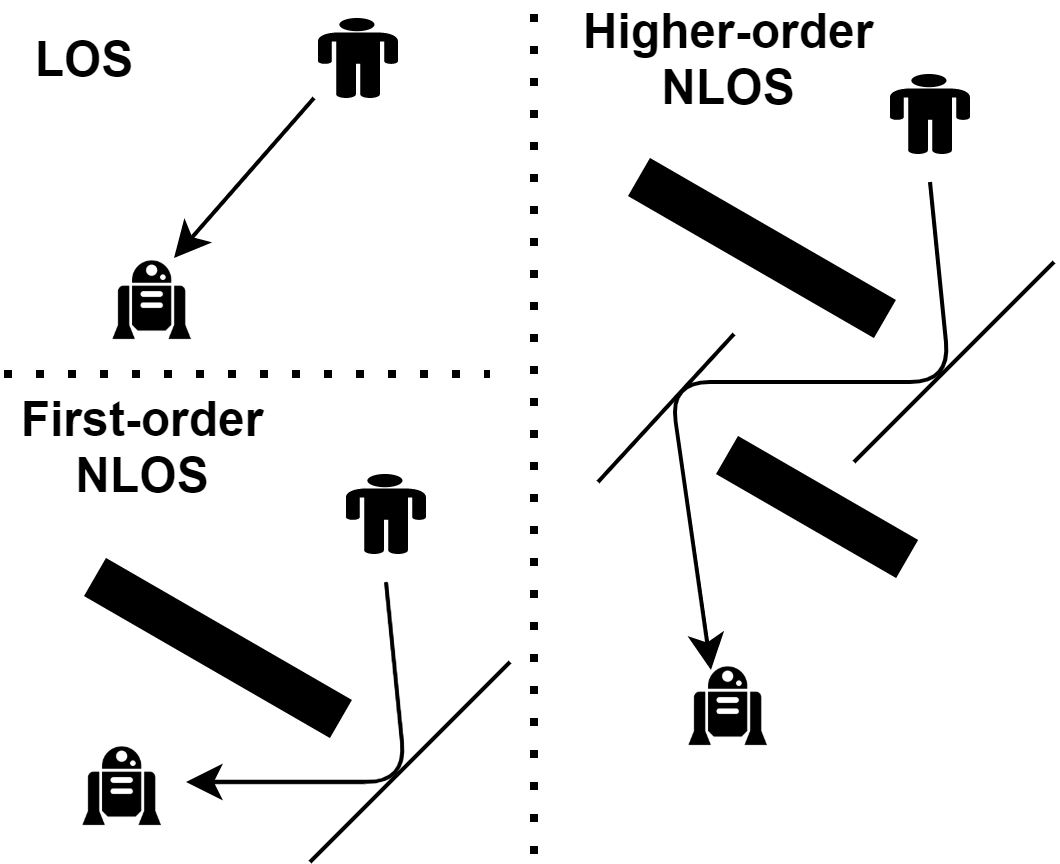}
  \caption{A demonstration of the LOS, First-order NLOS,
  and Higher-order NLOS path.}
  \label{fig:link-state}
\end{figure}

Now, each sufficiently strong 
path is either LOS or NLOS where any NLOS
path arrives with one or more \emph{interactions}
such as diffractions, reflections, or transmissions.
As shown in Fig. \ref{fig:link-state},
we will classify the link as being in
one of four states
on the basis of the strongest received path:
\begin{itemize}
    \item \emph{LOS}:  The strongest received path
    is above the minimum threshold and is LOS;
    \item \emph{First-order NLOS:} At least one of received paths that are above the minimum threshold is NLOS
    with one interaction;
    \item \emph{Higher-order NLOS:}
    All sufficiently strong 
    paths from the TX to RX are NLOS
    with two or more interactions.
    \item \emph{Outage:}  There are no 
    sufficiently strong paths above the minimum threshold.
\end{itemize}
Given this classification, we consider two problems:
\begin{enumerate}
    \item \emph{Path estimation:}
    Estimate the parameters (angles of arrival,
    angle of departure, relative delay and 
    received SNR) for the strongest paths; and
    \item \emph{Link state classification:}
    Determine if the strongest path
    is LOS, First-order NLOS or Higher-order NLOS,
    or there is no sufficiently strong path.
\end{enumerate}

As we will see below, the reason we are interested
in this problem is that the angle of arrival of
LOS and first-order NLOS path 
have strong correlation with
the optimal direction for navigation.
Hence, if can reliably detect the link state
and estimate the angle of arrival of the
strongest path,
we can build a navigation system
that simply follows the estimated
strongest path angle of arrival.  

\subsection{Signaling and Array Modeling}
We wish to realistically model the 
measurement of the signals from which the
path estimation and link classification will be
performed.
To this end, we make the following assumptions:

\paragraph*{Array Modeling} As discussed in the introduction,
multiple arrays are critical in the mmWave range to provide
360 degree coverage \cite{raghavan2019statistical,xia2020multi}.  
To this end, 
we assume that the TX and RX have, respectively,
$N_{\rm arr}^{\rm tx}$ and $N_{\rm arr}^{\rm rx}$ 
mmWave arrays.
We let $N_{\rm ant}^{\rm tx}$ and $N_{\rm ant}^{\rm rx}$ denote the 
number of antennas in each array.
Since each array is typically designed to cover some angular region, we
will sometimes refer to an array as a \emph{sector}.

Although our simulations below consider
only 2D localization, the methods are general
and apply to 3D modeling as well.  
For 3D problems, we will use the notation 
$\Omega=(\phi,\theta)$ to denote an
azimuth and elevation angle pair.
In the case of 2D problem, $\Omega = \phi$
is the azimuth angle only with the elevation
angle being ignored.

Now, for each potential 
RX angle, $\Omega^{\rm rx}$, we let $\nbu_{\rm rx}(\Omega^{\rm rx})$ denote the array signature
across all the $N_{\rm arr}^{\rm rx}$ arrays.
In general, a receive array signature represents
the complex gains that would appear on a 
plane wave arriving from a given angle
\cite{heath2018foundations}.
For the case of multiple arrays, 
we can write the received-side
array signature as
\begin{equation} \label{eq:urxarr}
    \nbu_{\rm rx}(\Omega^{\rm rx})
    = \left[ 
    \nbu_{\rm rx}^{(1)}(\Omega^{\rm rx}), \cdots,
    \nbu_{\rm rx}^{(N_{\rm arr}^{\rm rx})}(\Omega^{\rm rx}) \right],
\end{equation}
where $\nbu_{\rm rx}^{(j)}(\Omega^{\rm rx})$ is 
spatial signature from the $j$-th array at the RX.
To model the arrays accurately, 
we will assume $\nbu_{\rm rx}^{(j)}(\Omega^{\rm rx})$ includes both the array gain and element gain.
Also, to account for mutual coupling
in densely-spaced arrays, we can apply
the coupling matrix normalization as described in \cite{kelley1993array}.
We will also assume the array spatial
signatures are normalized such that
$\|\nbu_{\rm rx}^{(j)}(\Omega^{\rm rx})\|^2$
is the directivity (in linear scale) of
array $j$ in the direction $\Omega^{\rm rx}$.
At the TX side, we define
$\nbu_{\rm tx}(\Omega^{\rm tx})$ similarly:
\begin{equation} \label{eq:utxarr}
    \nbu_{\rm tx}(\Omega^{\rm tx})
    = \left[ 
    \nbu_{\rm tx}^{(1)}(\Omega^{\rm tx}), \cdots,
    \nbu_{\rm tx}^{(N_{\rm arr}^{\rm tx})}(\Omega^{\rm tx}) \right].
\end{equation}

Finally, to consider the response on a single array,
we introduce the following notation:  Given
a receive angle $\Omega^{\rm rx}$, let
\begin{equation} \label{eq:urxhat}
    \widehat{\nbu}_{\rm rx}(\Omega^{\rm rx})
    = \left[ 
    0, \cdots, 0,
    \nbu_{\rm rx}^{(j)}(\Omega^{\rm rx}), 0, \cdots, 0 \right],
\end{equation}
with $j = \argmax \|\nbu_{\rm rx}^{(j)}(\Omega^{\rm rx})\|^2$.  Since $\|\nbu_{\rm rx}^{(j)}(\Omega^{\rm rx})\|^2$ is the directivity of the array $j$,
$\widehat{\nbu}_{\rm rx}(\Omega^{\rm rx})$ in \eqref{eq:urxhat} represents the array response
on the array with the highest gain for the
angle of arrival $\Omega^{\rm rx}$.
Similarly, we define
\begin{equation} \label{eq:utxhat}
    \widehat{\nbu}_{\rm tx}(\Omega^{\rm tx})
    = \left[ 
    0, \cdots, 0,
    \nbu_{\rm tx}^{(j)}(\Omega^{\rm tx}), 0, \cdots, 0 \right],
\end{equation}
where $j = \argmax \|\nbu_{\rm tx}^{(j)}(\Omega^{\rm rx})\|^2$.

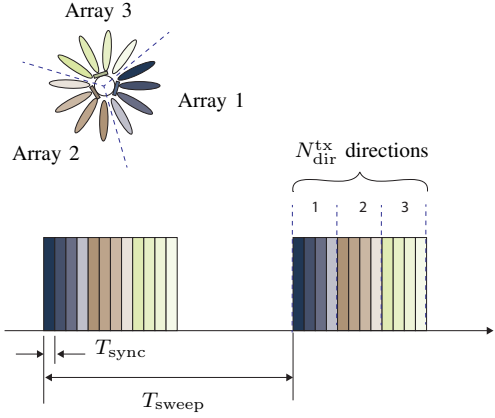
\begin{figure}
\footnotesize
\centering
    
\begin{tikzpicture}
    \node[inner sep=0] (image) at (0,0) {\includegraphics[width=6.5cm]{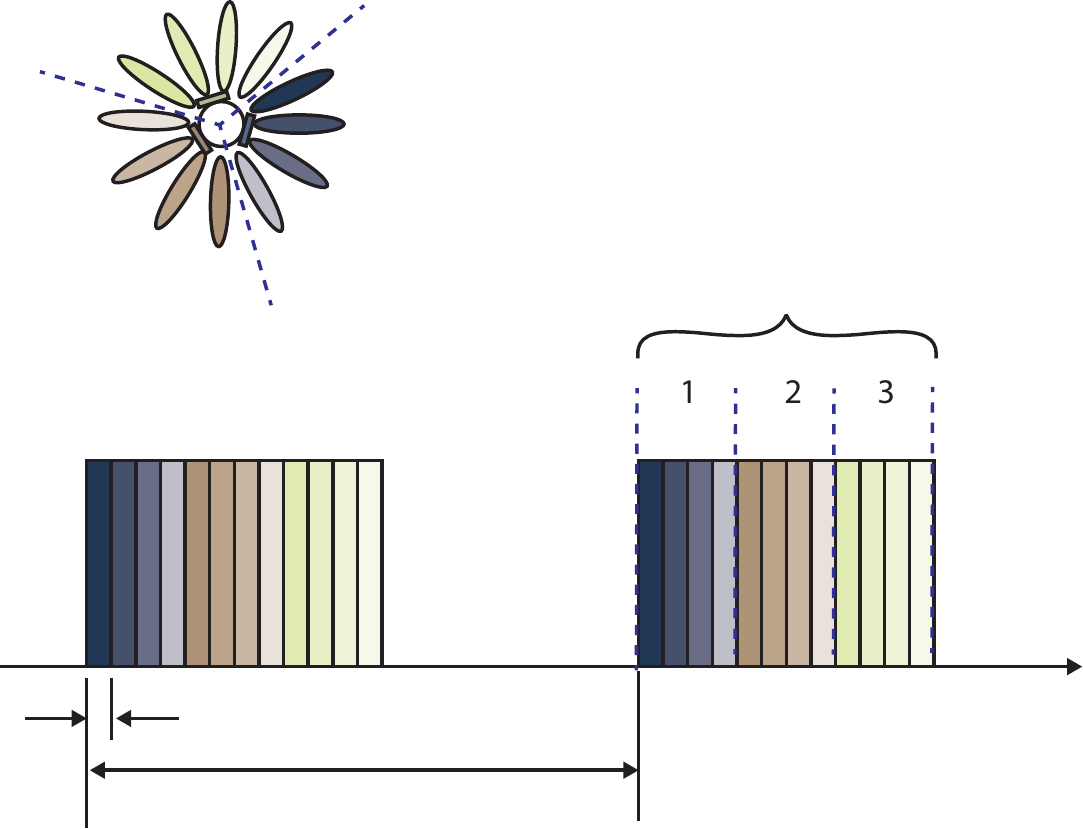}};
    \node[above=0 of image,yshift=-1.2cm,xshift=-0.5cm]
        {Array 1};
    \node[above=0 of image,yshift=-1.9cm,xshift=-2.7cm]        {Array 2};        
    \node[above=0 of image,yshift=0cm,xshift=-2cm]
        {Array 3};        
    \node [above=0 of image,yshift=-1.9cm,xshift=1.5cm]
        {$N_{\rm dir}^{\rm tx}$ directions};    
    \node [above=0 of image,yshift=-4.5cm,xshift=-1.7cm]
        {$T_{\rm sync}$};    
    \node [above=0 of image,yshift=-5.2cm,xshift=-1cm]
        {$T_{\rm sweep}$};    
 \end{tikzpicture}    
 
 \vspace{0.1cm}
   
    \caption{Example TX beam sweeping with $N_{\rm arr}^{\rm tx}=3$ arrays and 4 directions per array
    for a total of $N_{\rm dir}^{\rm tx}=12$ beam 
    directions.  The synchronization signals are
    sent once in each direction with the pattern
    repeating every $T_{\rm sweep}$ seconds.}
    \label{fig:beamsweep}
\end{figure}

\paragraph*{TX codebook}  
To enable detection, we assume the TX transmits a known
synchronization signal, sweeping through a sequence of directions
from the different TX arrays.  Similar to
\eqref{eq:utxarr}, we can represent a TX beamforming
vector with $N_{\rm arr}^{\rm tx}$ sub-vectors
with $N_{\rm ant}^{\rm tx}$ components each
representing the complex phases applied
to the antenna elements.
We will 
let $N_{\rm dir}^{\rm tx}$ denote the total number of 
TX directions.  For each TX direction $k$, we will let
 $\nbw_k^{\rm tx}$ denote the TX beamforming vector
 for that direction.  We will call 
 $\nbw_k^{\rm tx}$ 
a \emph{TX codeword}, the set of the 
TX codewords will be called the (multi-array) \emph{codebook}.

Several methods can be used for 3D codebook design
for mmWave systems
(e.g.\ \cite{song2015codebook,xia2020multi}).
In this work, we will simply assume the $k$-th 
TX codeword is of the form
\begin{equation} \label{eq:wtxang}
    \nbw^{\rm tx}_k = \frac{1}{\|\widehat{\nbu}_{\rm tx}(\Omegabar^{\rm tx}_k)\|}\widehat{\nbu}_{\rm tx}(\Omegabar^{\rm tx}_k),
\end{equation}
where $\Omegabar^{\rm tx}_k$ is some angle of departure
for the codeword and
$\widehat{\nbu}_{\rm tx}(\Omega^{\rm tx}_k)$ is as defined in \eqref{eq:utxhat}.  That is,
$\nbw^{\rm tx}_k$ is aligned to the TX steering vector for some TX angle $\Omegabar^{\rm tx}_k$.
We have used the overline in the transmit angle $\Omegabar^{\rm tx}_k$
to differentiate the TX angle from the 
angles of the paths $\Omega^{\rm tx}_\ell$
that will be described below.
The set of TX angles is the set
\begin{equation} \label{eq:txang}
    \left\{ \Omegabar^{\rm tx}_k, ~
    k = 1,\ldots,N^{\rm tx}_{\rm dir} \right\}.
\end{equation}
Observe that from \eqref{eq:utxhat}, 
each TX codeword is transmitted on only one
TX array at a time both to conserve power and not assume any local oscillator synchronization across TX arrays.

\paragraph*{TX beam sweeping}  
In each of the $N_{\rm dir}^{\rm tx}$ directions,
we assume 
the TX sends a known synchronization 
or reference signal
of duration $T_{\rm sync}$.
Hence, the total time to transmit all
$N_{\rm dir}^{\rm tx}$ directions 
is $T_{\rm sync}N_{\rm dir}^{\rm tx}$.
We assume the beam sweep 
pattern is repeated every $T_{\rm sweep}$ seconds with 
$T_{\rm sweep} \geq N_{\rm dir}^{\rm tx}T_{\rm sync}$.

An example is illustrated in Fig.~\ref{fig:beamsweep}
with $N_{\rm arr}^{\rm tx}=3$ TX arrays and four 
directions per array for a total of $N_{\rm dir}^{\rm tx}=12$
directions.

\paragraph*{Wireless channel model}
Let $x_k(t)$ denote the complex baseband
synchronization signal sent in the $k$-th
direction, $k=1,\ldots,N_{\rm dir}^{\rm tx}$.
We assume a standard multi-path cluster model
\cite{heath2018foundations} where the corresponding received signal is given by
\begin{equation} \label{eq:rxsig}
    \nbr_k(t) = \sum_{\ell = 1}^L 
    g_\ell 
    \nbu_{\rm rx}(\Omega^{\rm rx}_\ell)
    \nbu_{\rm tx}(\Omega^{\rm tx}_\ell)^\top \nbw^{\rm tx}_k 
    x_k(t-\tau_\ell) + \nbv_k(t), 
\end{equation}
where  
$\nbw^{\rm tx}_k$ is the TX beamforming for
direction $k$ as given in \eqref{eq:wtxang},
$\nbr_k(t)$ is the complex baseband signal across
all $N_{\rm arr}^{\rm rx}N_{\rm ant}^{\rm rx}$
receive antenna elements,
and $\nbv_k(t)$ is AWGN noise that we assume 
i.i.d.\ across RX antennas.  In \eqref{eq:rxsig},
$L$ is the number of paths, and for each path $\ell$,
$\Omega_\ell^{\rm rx}$ and $\Omega_\ell^{\rm tx}$
are the RX and TX angles,
$\nbu_{\rm rx}(\Omega^{\rm rx}_\ell)$ and
$\nbu_{\rm tx}(\Omega^{\rm tx}_\ell)$ are the steering
vectors of the arrays at those angles,
$g_\ell$ is the complex path gain,
and $\tau_\ell$ is the path delay.  For simulation,
one can generate the path parameters via statistical models
such as \cite{3GPP38901}, although as we will discuss below,
in this work the path parameters will be found
from ray tracing.

%% file: algo.tex
\section{Proposed Algorithm for Path Estimation
and Link State Classification}
\label{sec:algo}

\subsection{Path Estimation
via Low-Rank Tensor Decomposition}

For the path estimation, we adapt a commonly-used
low rank tensor decomposition method
\cite{zhou2017low,wen2018tensor} with
modifications to handle the
multiple arrays and TX beam sweeping.
Define the spatial and temporal correlation 
\begin{equation} \label{eq:rxcorr}
    \rho_k(\tau,\Omega^{\rm rx})
    := \int \widehat{\nbu}_{\rm rx}(\Omega^{\rm rx})^*
    \nbr_k(t)x_k^*(t-\tau)\, dt,
\end{equation}
which is a complex function of the TX angle $k$,
RX angle $\Omega^{\rm rx}$ and delay $\tau$.
Under the channel model \eqref{eq:rxsig},
the signal will contribute one peak in the correlation
magnitude 
$|\rho_k(\tau,\Omega^{\rm rx})|^2$ for each path
$\ell$ when 
$(\tau,\Omega^{\rm rx})=(\tau_\ell,\Omega^{\rm rx}_\ell)$ when
the TX direction $\Omega^{\rm tx}_\ell$
is closely aligned the transmit angle $\Omegabar^{\rm tx}_k$.  Thus, in principle, we can locate the
paths from the peaks in the correlation function.

Following \cite{zhou2017low,wen2018tensor},
we use a low-rank tensor decomposition 
to extract these peaks.
Specifically, we first select 
a discrete set of delays and RX angles:
\begin{align} 
    &\tau = \taubar_i, \quad
    &i=1,\ldots,N_{\rm dly}, \\
    &\Omega^{\rm rx} = \Omegabar^{\rm rx}_j,
    \quad &j=1,\ldots,N^{\rm rx}_{\rm dir}, 
\end{align}
where $N_{\rm dly}$ is the number of 
delay hypotheses and $N_{\rm dir}^{\rm rx}$
is the number of RX angular directions.
Again, note that we use the overlines
in $\taubar_i$ and $\Omegabar^{\rm rx}_j$ to
differentiate them from the true delays
and angles $\tau_\ell$ and $\Omega^{\rm rx}_\ell$.
We then evaluate the complex correlation 
\eqref{eq:rxcorr} at the discrete values to 
obtain a third-order tensor:
\begin{equation}
    S[i,j,k] = \rho_k(\taubar_i, \Omegabar^{\rm rx}_j).
\end{equation}
As described in \cite{zhou2017low,wen2018tensor},
we then find an approximate low-rank decomposition,
\begin{equation}
    S[i,j,k] \approx \sum_{\ell=1}^R
        a_{\ell i} b_{\ell j} c_{\ell k},
\end{equation}
where $R$ is the estimated rank, and
 $\nba_\ell$, $\nbb_\ell$ and $\nbc_\ell$
 are the basis vectors for each rank one term
 in the delay, RX angle and TX angle dimensions.
Any low-rank tensor decomposition method can be used (see, e.g.\ \cite{grasedyck2013literature} for a survey).  In this work, we will use
two-way PCA by first performing PCA
flattening the $(j,k)$ dimensions and then
performing PCA on the $(j,k)$ matrix.

Following the low-rank tensor decomposition,
we can estimate for the 
delays, TX and RX angles can 
be found from the peaks in the basis vectors:
\begin{subequations}
\begin{align}
    &\widehat{\tau}'_\ell = \overline{\tau}_i, \quad
    &i = \argmax |a_{\ell i}| \\
    &\widehat{\Omega}^{\rm rx}_\ell = \overline{\Omega}^{\rm rx}_j, \quad
    &j = \argmax |b_{\ell j}| \\
    &\widehat{\Omega}^{\rm tx}_\ell = \overline{\Omega}^{\rm tx}_k, \quad
    &k = \argmax |c_{\ell k}|.
\end{align}
\end{subequations}

Since we do not assume timing synchronization 
between the TX and RX,
the estimated delays $\widehat{\tau}'_\ell$
are only meaningful as \emph{relative}
values.  For this reason, we convert the
raw delay estimates to relative delay 
estimates
\begin{equation} \label{eq:taurel}
    \widehat{\tau}_\ell =
    \widehat{\tau}'_\ell - \min_{\ell=1,\ldots,K}
    \widehat{\tau}'_\ell.
\end{equation}

We can also estimate the RX SNR
of the path as:
\begin{equation} \label{eq:gamest}
    \gamma_\ell := \frac{\|\nba_\ell\|^2\|\nbb_\ell\|^2\|\nbc\|^2_\ell}
    {E_{\rm avg}},
\end{equation}
where the numerator represents the
total energy along the estimated
$\ell$-th direction and the denominator
is the average energy per sample,
\[
    E_{\rm avg} := 
    \frac{1}{N_{\rm dly}N_{\rm dir}^{\rm tx}N_{\rm dir}^{\rm rx}}
    \sum_{ijk} |S[i,j,k]|^2.
\]
The resulting procedure thus produces 
a set of path estimates
\begin{equation} \label{eq:pathparam}
    \left\{
    (\tauhat_\ell, \Omegahat_\ell^{\rm rx},
    \Omegahat_\ell^{\rm tx},\gamma_\ell),
    ~ \ell = 1,\ldots, K \right\},
\end{equation}
that includes the relative delay,
RX and TX angles and SNRs of the path.
We will fix $K$ in this work since we 
have an SNR indication $\gamma_k$
to discard low energy paths.

\subsection{Configuration and Synchronization Issues}
\label{sec:sync}
The methods described above apply to any scenario where the target device
can transmit a mmWave beam sweeping synchronization signals.
Such signals are commonly used in mmWave RADAR and can be used in this application
if the target can carry a RADAR transmitter.
If the target device has a standard 5G mmWave device, then the existing
management and positioning signals may be used
\cite{giordani2018tutorial,keating2019overview,dwivedi2021positioning},
but some additional configuration capabilities may be needed.
For example, if the target is a 5G UE device, the path estimation could
potentially be performed from
the uplink sounding reference signals (SRS).
The mobile agent could then attempt to listen to the SRS signals for localizing the target.
Of course, normally the SRS are directed to the serving base station.
Hence, 
for the signaling method described above, the UE would need to receive some configuration
from the network to transmit the SRS in a sweeping pattern.  
Also, if the UE does not have a connection to a mmWave
base station (a likely scenario since the target may be inside where the mmWave
signals are blocked), the UE would need to be configured over a sub-6 GHz
carrier and be permitted to transmit even in absence of a detected network.

Whichever signals are used, one issue is the transmit synchronization.
For the mobile agent RX to know the transmit angle, $\widehat{\Omega}^{\rm tx}_\ell$
in \eqref{eq:pathparam}, the RX would need to know which beams are being transmitted
at which times.  In addition, the TX would need to inform the RX of the 
TX directions.  In the cellular case, this information would also need
to be conveyed in the configuration.  Since this may not always be possible,
we investigate in the simulations below the cases when the TX angles is and is not available.

\subsection{Link State Classification
Neural Network}
\label{sec:linkstatenn}

The next step is to classify the link state
(LOS, First-Order NLOS, Higher-Order NLOS
or Outage) from the parameter estimates \eqref{eq:pathparam}
of the detected paths.
We use the estimates from a fixed number, $K=5$, paths, 
taking the paths with the \emph{strongest} estimates SNRs.
In general, increasing
the number of paths increases the number of features in our model and hence can result in
over-fitting.  We found $K=5$ to provide
the lowest generalization error.
Note that we take the 
$K=5$ \emph{strongest} paths, not the $K$
\emph{earliest} paths.  
Hence, we do not
miss any strong paths that arrive late.

Since the data \eqref{eq:pathparam}
are heterogeneous, we then transform the inputs to a common scale -- a critical 
step to use neural networks.
Although the methods can be applied for 3D angles,
for the remainder of the paper, we will focus on 2D problems
that only use the azimuth angles.  That is, the angle estimate $\widehat{\Omega}^{\rm tx}_\ell$ 
$\widehat{\Omega}^{\rm rx}_\ell$ are scalar
azimuth angles.
These angles are in the range $[-\pi,\pi]$, and are scaled from $[-1,1]$.
The SNR parameter $\gamma_\ell$ are transformed to 
scaled value in $z_\ell \in [0,1]$ with the transformation
\begin{equation} \label{eq:gamscale}
    z_\ell = \max\left\{ 0, 
    \min\{ 1, \frac{\gamma_\ell - \gamma_{\rm min}}{\gamma_{\rm max} - \gamma_{\rm min}}\} \right\},
\end{equation}
for some fixed parameters $\gamma_{\rm min}$ and $\gamma_{\rm max}$.
Under the scaling \eqref{eq:gamscale}, paths weaker than the threshold $\gamma_{\rm min}$
result in $z_\ell = 0$ and very strong paths ($\gamma_\ell \geq \gamma^{\rm max}$)
are mapped to $z_\ell  = 1$.  
We used fixed limits $\gamma_{\rm min}=5$\, \si{dB} and $\gamma_{\rm max}=50$\, \si{dB}.
Also, when less than $K$ paths are detected, we set the corresponding scaled SNR values 
to zero to indicate the path does not exist.

For the delays, we measure the values relative to the first detected path \eqref{eq:taurel}.
Subtracting the first path is required since we do not assume absolute timing.
We then scale the relative delays by a fixed value of \SI{100}{ns} which corresponds to
\SI{30}{m} of propagation distance.

In total, the pre-processing results in scaled values of the four parameters 
of the each paths \eqref{eq:pathparam}.  With data from $K=5$ paths, the neural network takes
as an input $4K=20$ values.   If the TX angles $\widehat{\Omega}^{\rm tx}_\ell$
are not assumed to be available, then the neural network will have $3$ values per path
with a total of $5(3)=15$ values.  As we will describe below, 
we will then apply a simple 2 layers
fully connected network.  
The output is a four-way softmax for the four link states,
LOS, First order NLOS, High order NLOS and Outage.

%% file: raytracing.tex
\section{Creation of the 
Robotic and Wireless Dataset}
\label{sec:raytracing}

We validate the methods on the
AI Habitat robotic simulation environment \cite{savva2019habitat} with the Gibson 3D indoor models \cite{xia2018gibson}. 
Gibson’s underlying database of spaces includes 572 full buildings composed of 1447 floors covering a total area of  \SI{211000}{\meter\squared}.  The
indoor environments include
complete 3D camera and LiDAR data,
and maps from actual interior environments.
These 3D models can be imported into the
AI Habitat simulation that includes
a complete kinematic model of the robots
The simulation environment has been widely-used
for validating indoor SLAM and 
navigation algorithms, see e.g.\
\cite{yu2020meta,xia2020interactive}.

One key contribution of this work is to 
augment this dataset with a wireless coverage maps.  This will create what we believe
is the first dataset with camera, LiDAR
and wireless data integrated into a robotic 
simulation environment.  
In addition to validating the
algorithms in this work,
the dataset can be used for other
positioning work as well as cloud robotics
experiments with wireless connectivity.

\begin{figure}
  \centering
  \includegraphics[width=0.6\linewidth]{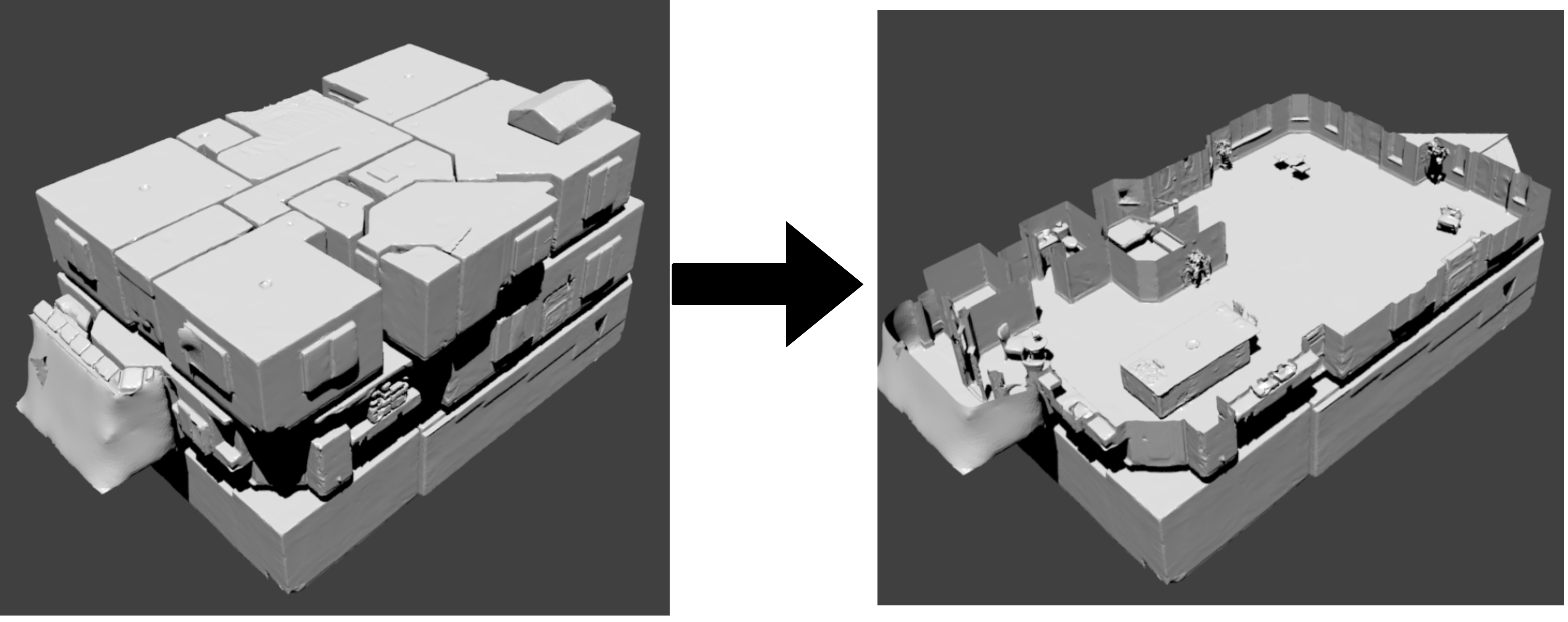}
  \caption{Most 3D models in the Gibson data set are of multi-storey residences. In this example, the left figure
  shows a three storey house. We exact the second floor,
  as shown on the right,
  to generate our test layout.}
  \label{fig:pick_one_floor}
\end{figure}

To capture the wireless coverage,
we use the powerful ray tracing package, Wireless InSite by Remcom \cite{Remcom}
which has also been used in several other
recent mmWave studies such as \cite{taha2021millimeter, khawaja2017uav, xia2020generative}.
Ray tracing uses a high-frequency approximation to simulate the electromagnetic
paths between any two locations.
As the Gibson dataset contains highly detailed and fine-grained 3D models, we need to pre-process them in order to reduce the computational effort when we use them as ray tracing environments.  
Since we 
consider the navigation and positioning problems in 2D and most of the buildings in the data set are multi-story residence buildings, we first select and extract one floor from each
-- an example is shown in Fig.~\ref{fig:pick_one_floor}. 

\begin{figure}
  \centering
  \includegraphics[width=0.6\linewidth]{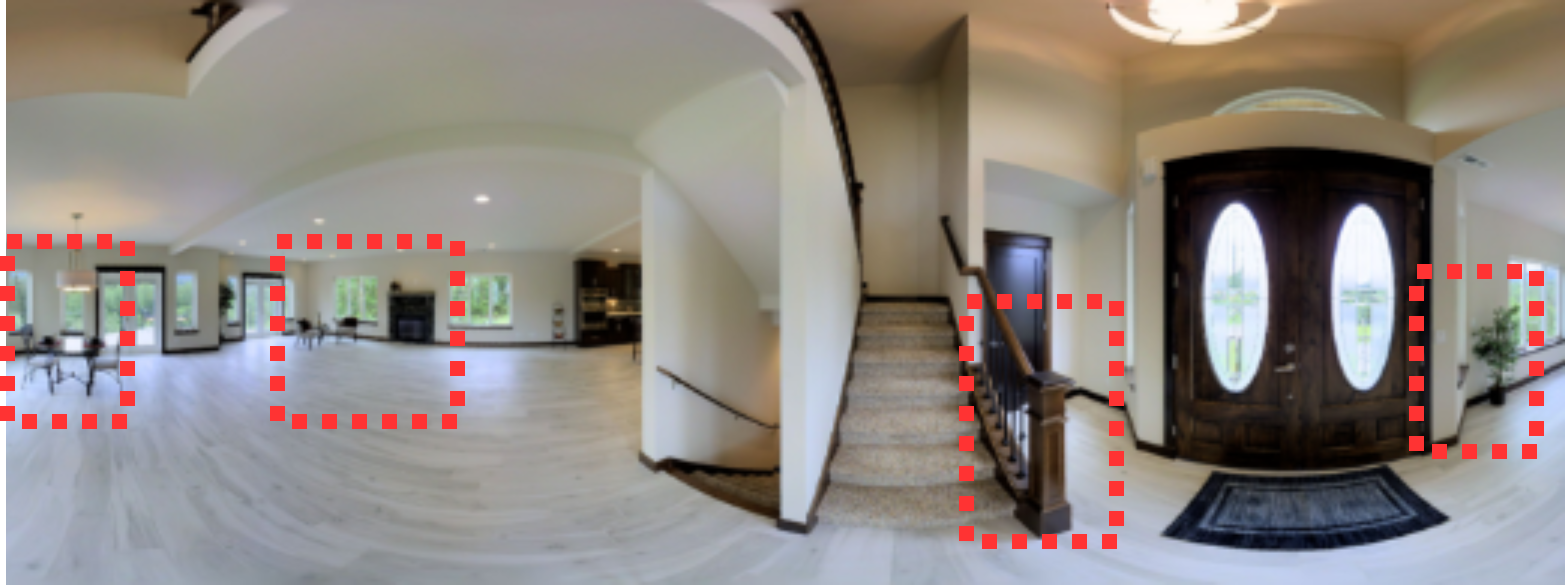}
  \caption{A typical indoor environment in the Gibson data set \cite{xia2018gibson} with
  furniture,
  plants, and irregular building structures that significantly increases computational load for ray tracing.}
  \label{fig:room-inside}
\end{figure}

\begin{figure}
  \centering
  \includegraphics[width=0.8\linewidth]{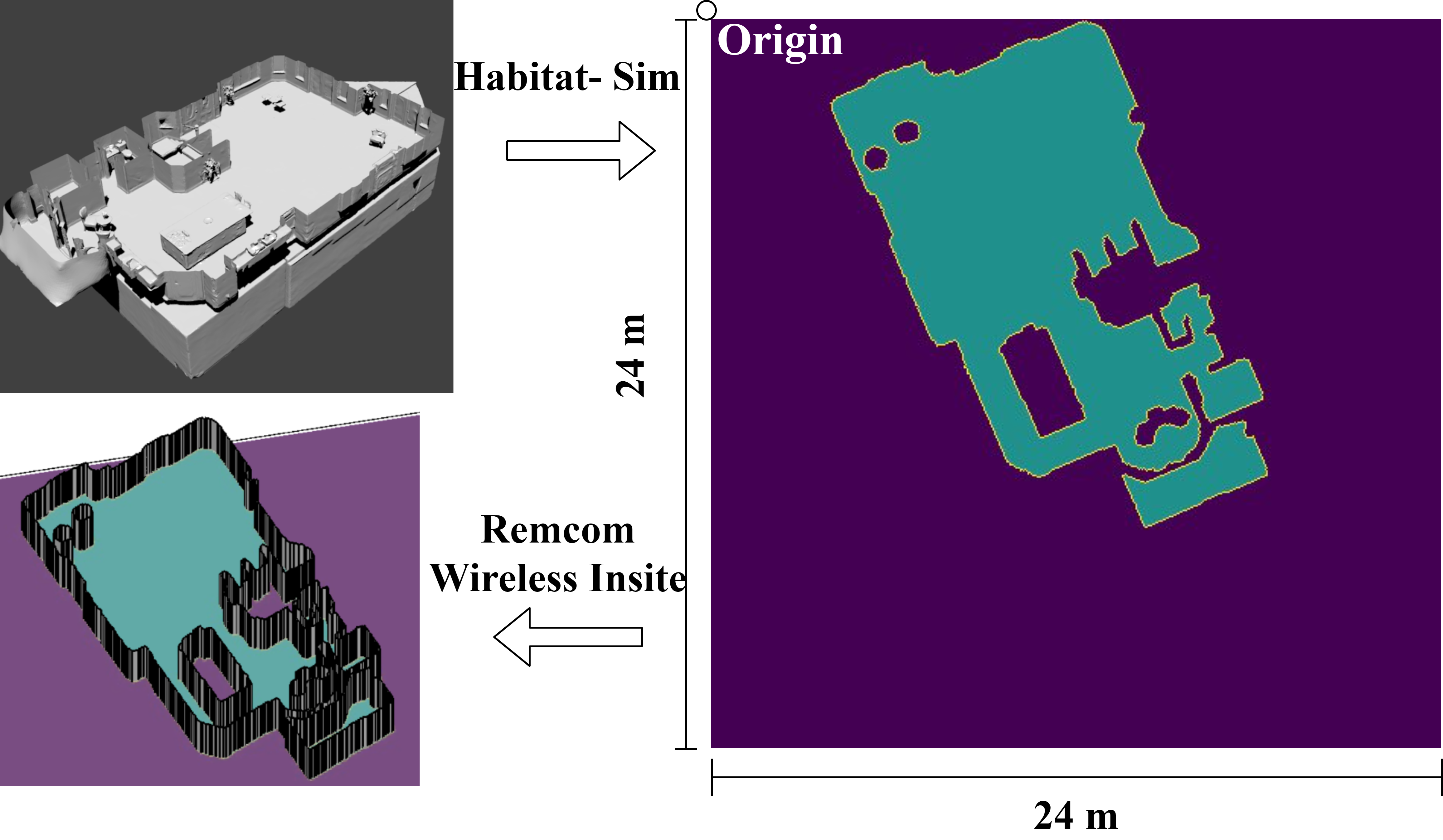}
  \caption{An example of a 24$\times$24 meter indoor environment map generated by the
  Habitat-Sim \cite{habitat19iccv}. The Wireless InSite tool by Remcom
  \cite{Remcom} uses the top-down map to rebuild a
  3D indoor model. Some assumptions are made to reduce the high computational costs of ray tracing.}
  \label{fig:map_convert}
\end{figure}

Next, as shown in Fig.~\ref{fig:room-inside}, the indoor environment contains furniture, plants, and irregularly constructed structures. 
These objects often have many non-smooth reflecting surfaces, which significantly increases the amount of computing required for the ray tracing. 
To simplify the 3D model, we utilize Habitat-Sim \cite{habitat19iccv, szot2021habitat} to capture a floor plan
-- making the model essentially 2D.
Also, as far as possible, the irregular and uneven walls, ceilings, and floors of the original 3D model are transformed into smooth planes.  The edges of the 
simplified 2D floor plan
are then vertically extended to create walls.
The resulting 3D map can then be imported to
Remcom Wireless Insite
-- See ~Fig.~\ref{fig:map_convert}.

\begin{figure}
  \centering
  \includegraphics[width=0.7\linewidth]{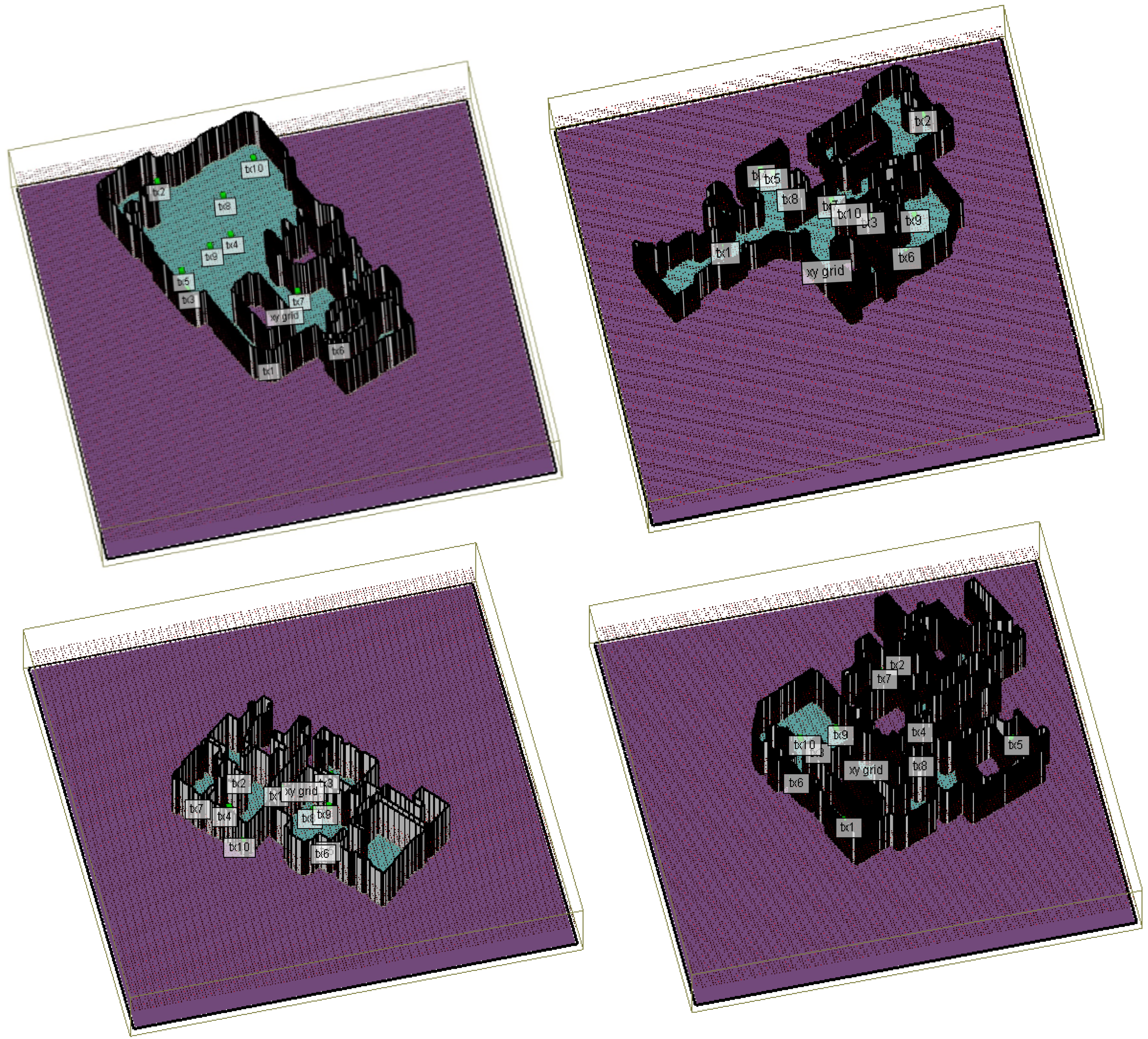}
  \caption{Four examples of ray tracing simulation area. In each map, 
  green dots represent transmitter locations and the red dots   represent the receiver grid with total 25,600 receivers. }
  \label{fig:map_demostration}
\end{figure}

For each imported map, we then place
the transmitters in 10 randomly selected locations
representing 10 possible target positions.  
The ray tracing is then used to estimate the wireless paths at RX locations
on a 160 $\times$ 160 grid with 
$0.15 \times 0.15$\, \si{m} grid
representing
a total area of \SI{24}{\meter\squared}.
Example ray tracing simulations areas
are shown in Fig.~\ref{fig:map_demostration}.

The ray tracing is performed at \SI{28}{GHz},
the most commonly-used frequency for 5G mmWave
devices~\cite{dahlman20205g}. We ignore the difference of material and
treat all walls as the ITU (International Telecommunication Union) layered drywall whose permittivity is 2.94 (\si{F/m}) and conductivity is 0.1226 (\si{S/m}) in \SI{28}{GHz}.


\begin{figure}
  \centering
  \includegraphics[width=0.8\linewidth]{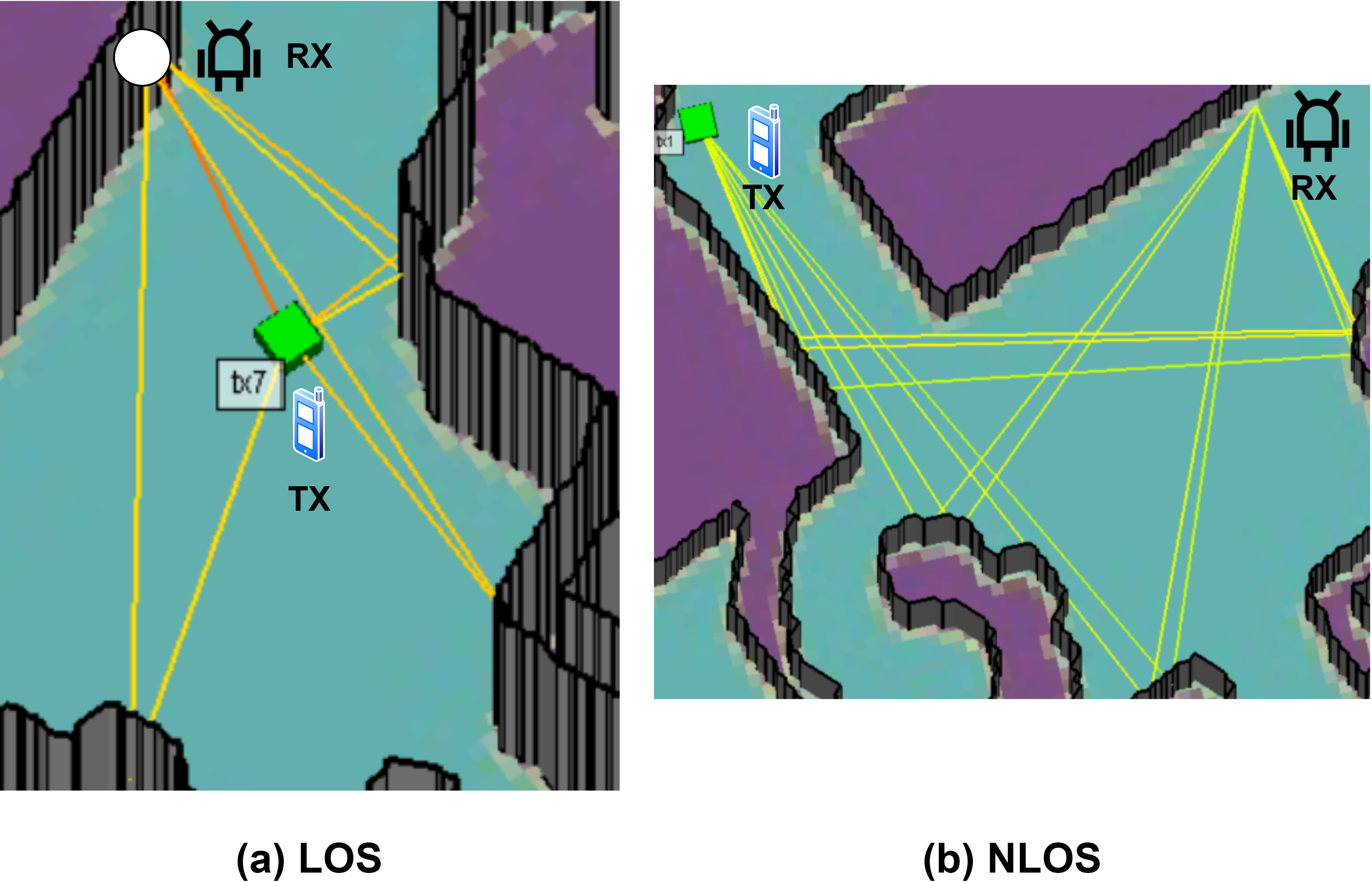}
  \caption{LOS (left) 
    and NLOS (right) examples of the ray tracing paths. (Some weaker paths are not shown).
  }
  \label{fig:ray_tracing_example}
\end{figure}

For each TX-RX location,
the ray tracing produces an estimate of the
set of large scale parameters 
\[
    (g_\ell, \Omega^{\rm rx}_\ell, 
    \Omega^{\rm tx}_\ell, \tau_\ell),
    \quad 
    \ell=1,\ldots,L,
\]
for the paths in the channel model
\eqref{eq:rxsig}.  Example
ray tracing outputs are shown in Fig.~\ref{fig:ray_tracing_example}.
Using the true signal parameters,
we can then simulate the path estimation
algorithms in Section~\ref{sec:algo}
to create estimates of the 
path parameters \eqref{eq:pathparam}.
The specific parameters for the arrays
and transmitter are discussed in the 
next section.

In total, the procedure is run on 38 unique environments from the Gibson dataset
with 10 TX locations in each environment.  For each environment and
TX location, the ray tracing results
in a map of the true 
wireless paths at each RX location.
The path estimation simulation creates
a second map of the estimated paths
at each RX locations.  Combined
with the original Gibson data,
we have thus created a unified dataset
where 3D models with camera and LiDAR
data are augmented with wireless ray tracing
and wireless path estimation.
The dataset is made available in \cite{mmwRobotNav}.


%% file: sim.tex
\section{Path Estimation and Link State Classification Simulation Results}
\label{sec:sim}

In this section, we describe the simulation parameters and measured performance of
the path estimation and link state classifier
on the data set in Section~\ref{sec:raytracing}.
We will discuss the integration with 
navigation in the subsequent section.

\subsection{Array Modeling}
\begin{figure}[ht]
  \centering
  \includegraphics[width=1.0\linewidth]{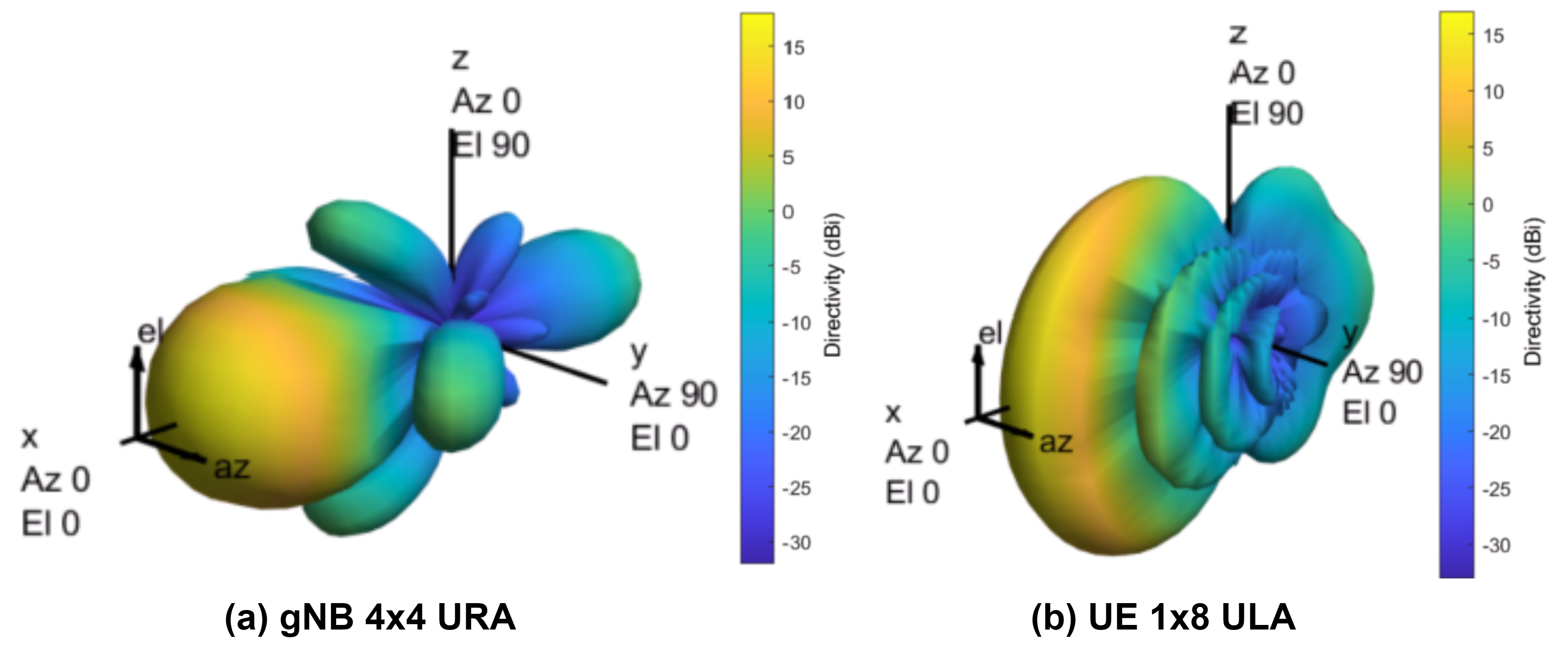}
  \caption{The pattern of a gNB antenna array 
  and a UE antenna array. Arrays are aligned 
  so that its bore-sight is on the x-axis.
  An antenna's directivity is a component of its gain.
  When the target is located in the main lobe (yellow areas in the figure), the maximum antenna gain is received.}
  \label{fig:ant_array_pattern}
\end{figure}

We assumed a \SI{28}{GHz} 
array similar to several
recent published 5G designs and layouts
\cite{raghavan2019statistical,gu2021development}. The parameters for the arrays and signaling are shown in Table.~\ref{table:sim_ant_para}.
The target TX is assumed to be a UE device
and the RX is a gNB.  The arrays use
microstrip patch antennas and the beam
patterns of each array when directed at boresight are shown in Fig.~\ref{fig:ant_array_pattern}.

\begin{figure}[ht]
  \centering
  \includegraphics[width=0.6\linewidth]{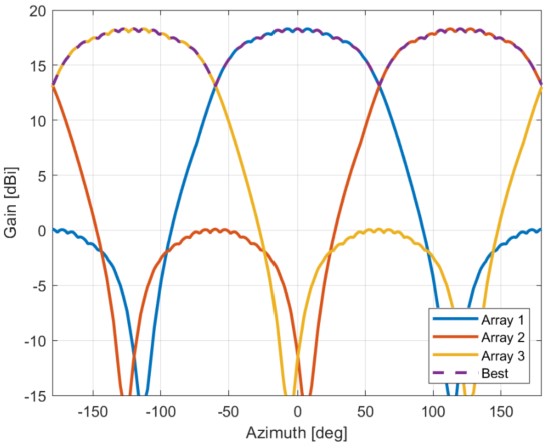}
  \caption{Array gain including the
  element gain from each gNB array 
    as well as the best for all three arrays.  We see that by using multiple arrays
    we can obtain full azimuth coverage.}
  \label{fig:array_coverage}
\end{figure}

As in
\cite{raghavan2019statistical},
we create the multiple antenna arrays at both the gNB and UE.   Specifically, we assume 
three arrays with azimuth angles
$0\degree$, $120\degree$, and $-120\degree$ and $0\degree$ elevation.
The multi-sector layout provides $360\degree$
coverage.  For example, 
Fig.~\ref{fig:array_coverage} plots the
maximum gain for each individual array and the maximum gain over all three arrays over $[-180\degree,180\degree]$ the azimuth directions.  We see that the minimum
gain is only approximately \SI{3}{dB} below
the maximum.  

For the beam search 
codebook at the gNB, we simply use the
beams on 48 angles uniformly spaced
in $[-180\degree,180\degree]$ in the azimuth
direction.  Since the gNB has 16 elements
per array and three arrays, the use of 48 beams corresponds to the number of orthogonal
spatial degrees of freedom.  Similarly,
in the UE we use 24 beams.

\begin{table}[]
\caption{Path Estimation 
Simulation Parameters}
\label{table:sim_ant_para}
\begin{center}
\begin{tabular}{|l|l|l|}
\hline
\multirow{2}{*}{\textbf{Item}} & \multicolumn{2}{c|}{\textbf{Value}}                                  \\ \cline{2-3} 
                               & \multicolumn{1}{c|}{\textbf{UE}} & \multicolumn{1}{c|}{\textbf{gNB}} \\ \hline
                        
Carrier  & \multicolumn{2}{l|}{\SI{28}{GHz}}                        \\ \hline
Bandwidth  & \multicolumn{2}{l|}{\SI{400}{MHz}}  \\ \hline
Antenna element & \multicolumn{2}{l|}{Mircostrip patch antenna}                        \\ \hline
Number of Arrays & 3 \cite{8606243}   & 3           \\ \hline
Array Size   & 8 (1 $\times$ 8 ULA)             & 16 (4 $\times$ 4 UPA)             \\ \hline
Transmit Power                & \SI{23}{dBm} &
-  \\ \hline
Losses                         & \multicolumn{2}{l|}{\SI{6}{dB} including noise figures \cite{7956180}}                    \\ \hline
Number beams           & 24 & 48 \\ \hline
\end{tabular}
\end{center}
\end{table}


\subsection{Path Estimation}
As shown in Table~\ref{table:sim_ant_para},
the TX power was set to \SI{23}{dBm} 
with a sample rate of \SI{400}{MHz}
-- standard in 5G deployments~\cite{dahlman20205g}.
For the positioning signal, we transmitted 
a 2048 sample
random waveform for a total of
duration of \SI{5.1}{\micro\second} to
enable the signal to be transmitted entirely
in one 5G NR OFDM symbol.  The correlation
\eqref{eq:rxcorr} was performed via an FFT.
Since the positioning accuracy
is only dependent on the sample rate, duration and power
--- such as those used in 3GPP positioning 
\cite{3GPP38885,3GPP38857,3GPP21916,keating2019overview,dwivedi2021positioning} ---
would have resulted in similar performance.

\begin{figure}[ht]
  \centering
  \includegraphics[width=1\linewidth]{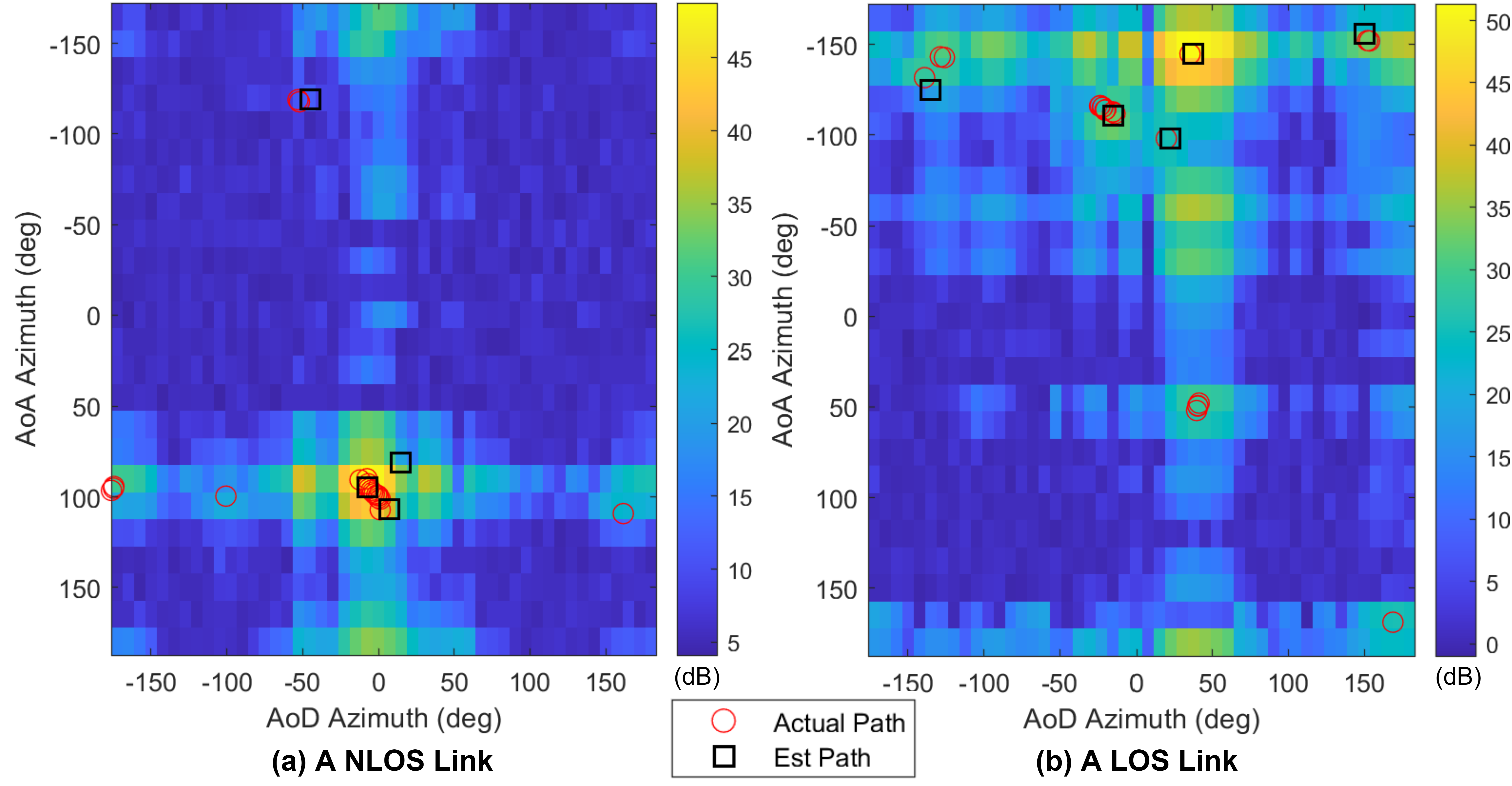}
  \caption{Example received power spectrum along TX-RX direction pair
    along with the location of the actual ray tracing 
    paths of a LOS link and a NLOS link.
    The black squares indicate the result of path 
    estimation by the low-rank tensor decomposition in Section~\ref{sec:algo}.}
  \label{fig:h_peak}
\end{figure}

The tensor decomposition path estimation algorithms in Section~\ref{sec:algo} were then run at each
location.  Fig.~\ref{fig:h_peak} depicts a 
result of the estimated paths at one LOS location
(left) and NLOS location (right).  The 
red circles represent the angular locations of true paths where you can clearly observe
path clusters from the scattered reflections
and diffractions.  The black squares are the
estimated path locations and generally extract
the cluster centers.

\begin{figure}[ht]
  \centering
  \includegraphics[width=0.7\linewidth]{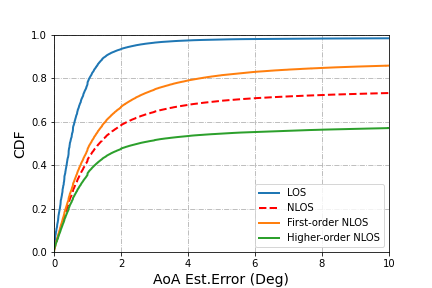}
  \caption{Distribution of the 
  absolute error between 
  the estimated strongest path's AoA  from 
  channel sounding and
  the AoA of the strongest path in real ray tracing data set.}
  \label{fig:aoa_est_err}
\end{figure}

The overall accuracy of the method is
depicted in Fig.~\ref{fig:aoa_est_err}
which shows the CDF of the  AoA error on 
the strongest path across 
the 38 environments and 10 transmitter
locations described in the dataset 
in Section~\ref{sec:raytracing}.
The CDF is plotted separately for LOS and NLOS
cases with the NLOS case plotted separately
for the case of first-order and higher-order NLOS.
We see that the strongest path's AoA errors differ greatly between the three link states.
Indeed, $98\%$ of LOS links 
and $85\%$ of first-order NLOS links have errors less than $5\degree$.
In contrast, for higher-order NLOS links, more than $40\%$ of errors are larger than  $10\degree$.
For this reason, the angular estimates for the
LOS and first-order
NLOS links are much more reliable when used
for navigation.  

\subsection{Link State Classification}
We next evaluated the link state neural network
described in Section~\ref{sec:linkstatenn}.
Our network attempts to classify the link
state (LOS, First-Order NLOS, Higher-Order NLOS,
and outage) from the estimated path parameters
\eqref{eq:pathparam}.
As inputs, we use the parameters from the 
$K=5$ \emph{strongest} detected paths.
The parameters of this link-state neural network are shown in Table \ref{table:link_state_nn_config}.
When the TX
When the TX angle is available, there are 4
parameters per path for a total of $4K=20$
inputs.  We use 
a simple fully connected neural network 
with two layers with 8 and 6 hidden units
with the ReLU activation \cite{nair2010rectified}.
We also compare our neural network method to a modified method from \cite{huang2020machine}, one of the best prior work algorithms.
In paper \cite{huang2020machine}, Huang and Molisch et al. propose the machine learning based LOS and NLOS identification methods for the MIMO system in vehicle-to-vehicle (V2V) networks. 
The method of 
\cite{huang2020machine} uses a set of custom features (static and time-varying characteristics),
while our proposed approach uses the raw MPC (the multipath component)
data.

The network is trained on 18 of the 38 environments and tested on the remaining 20.
As described in Section~\ref{sec:raytracing},
each environment has 10 TX locations.
We train and test only on the valid indoor
RX locations in each map.
The resulting total 
training set contains 483690 unique TX-RX links while the validation set contains 563810 unique TX-RX links.

We train this network with links from a set of training maps, resulting in a pre-trained link-state classification neural network. We tested the robot's navigation algorithms on a other set of different maps. It should be emphasized that the test maps are completely new environments relative to the robot, which also shows that our pre-trained link-state classification neural network can be extended to any unknown new indoor environment.

\begin{figure}[ht]
  \centering
  \includegraphics[width=0.7\linewidth]{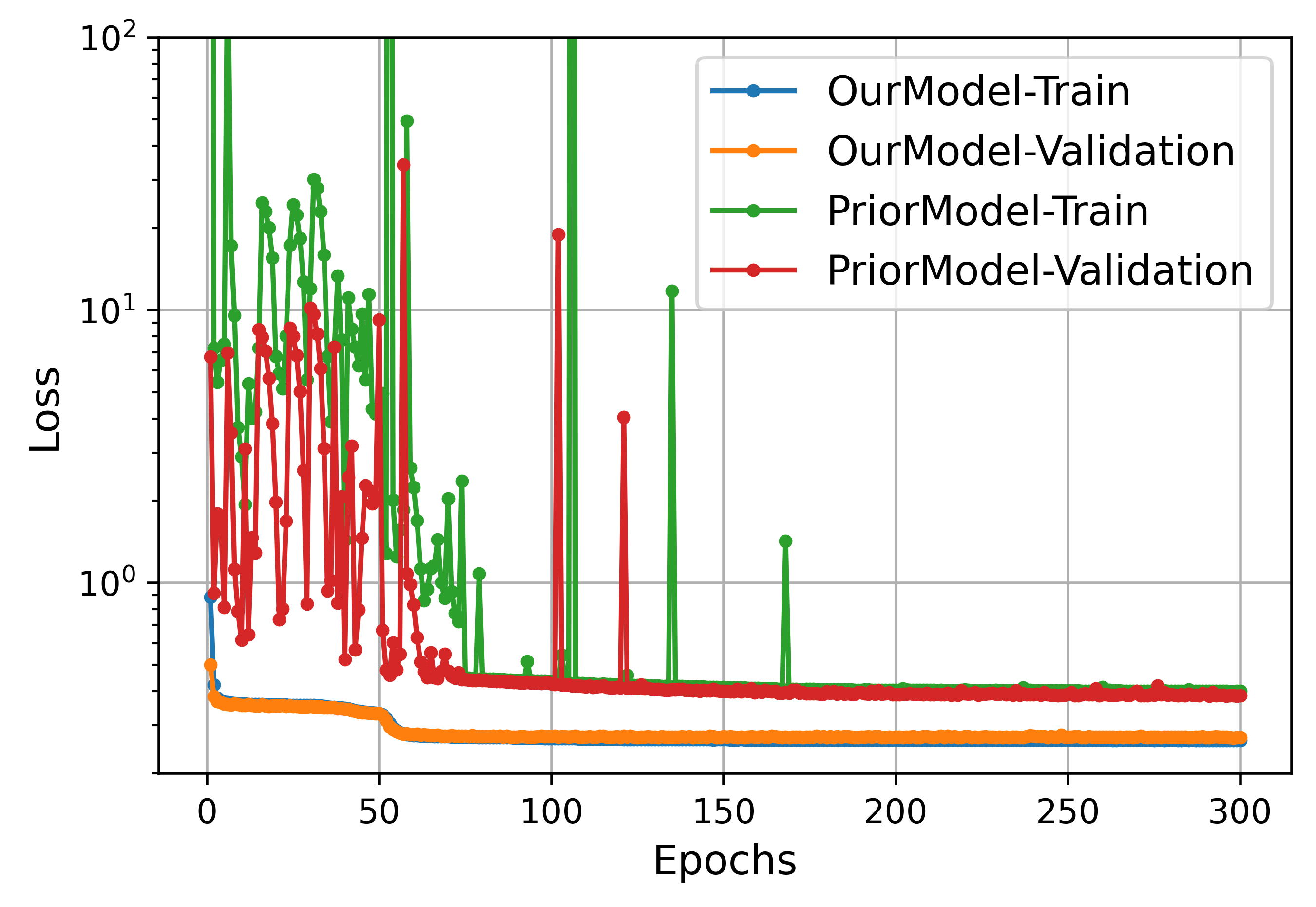}
  \caption{Loss of the link-state classification neural networks.}
  \label{fig:NN_loss}
\end{figure}

\begin{figure}[ht]
  \centering
  \includegraphics[width=0.8\linewidth]{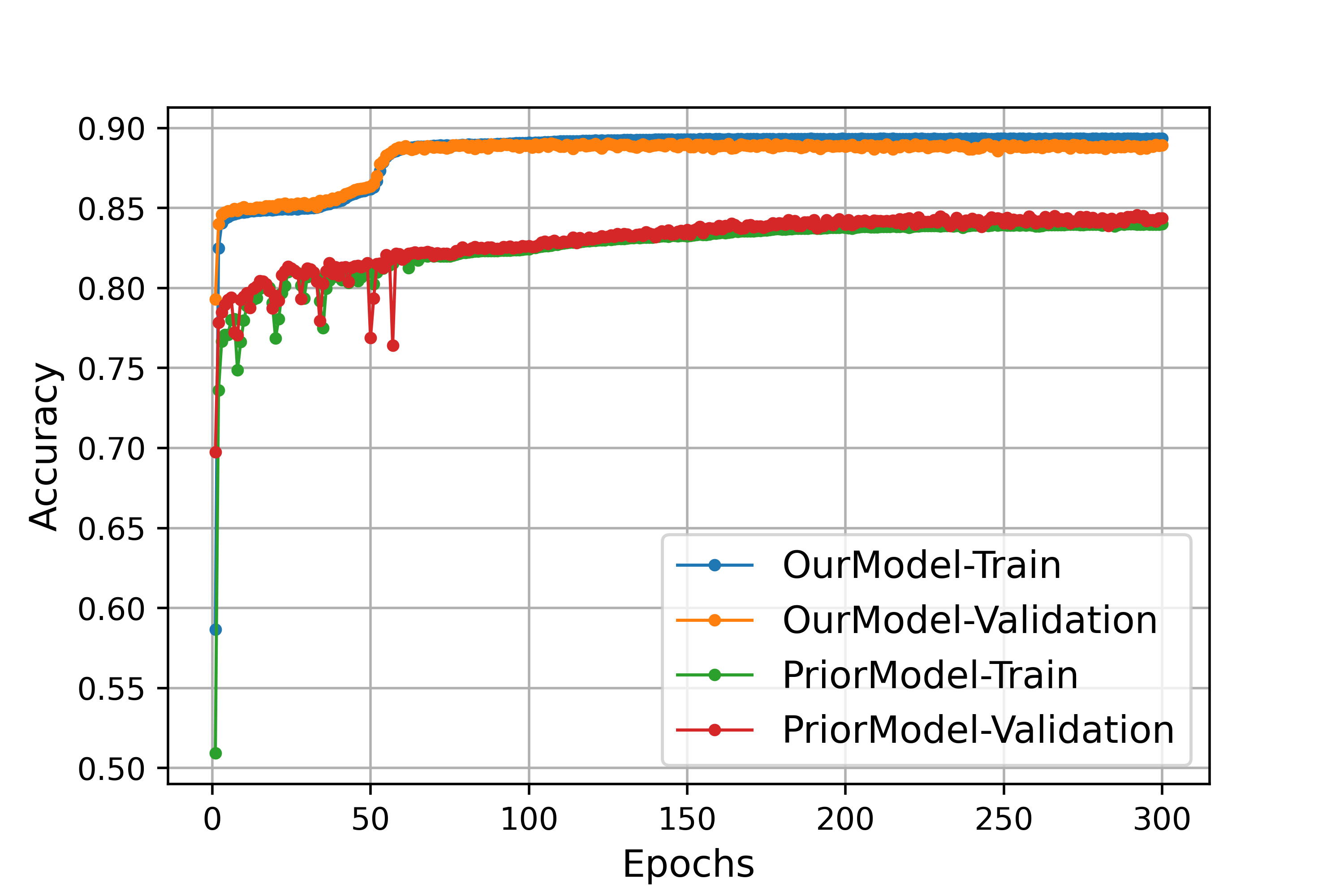}
  \caption{The 18 maps are used as the training set 
    and the other 20 maps are used as the test set.
    The yellow line shows the validation accuracy of our model which
    reached about $88\%$
    after about 100 epochs.}
  \label{fig:NN_acc}
\end{figure}

We use the Adam optimizer \cite{kingma2014adam}
with a learning rate of 0.001 for 300 epochs.
The loss is shown in Fig.~\ref{fig:NN_loss},
and the accuracy
is shown in Fig.~\ref{fig:NN_acc}.
The yellow line represents the accuracy of validation, while the blue line represents the accuracy of training.
From Fig.~\ref{fig:NN_acc}, we see that our model's validation accuracy reached 88\% after approximately 100 epochs,
while the modification of prior method obtains 84\%.
The main contribution of our network is considering first-order NLOS versus higher-order NLOS (a four-way link-state classifier), not this increase in accuracy.

\begin{table}[]
\caption{Link state neural network 
classifier configuration}
\label{table:link_state_nn_config}
\begin{center}
\begin{tabular}{l|l|}
\hline 
\multicolumn{1}{|l|}{\textbf{Parameter}} & 
\textbf{Value} \\ \hline
\multicolumn{1}{|l|}{Number of inputs}     & 20                                                                                    \\ \hline
\multicolumn{1}{|l|}{Hidden units}         & {[}8, 6{]}                                                                            \\ \hline
\multicolumn{1}{|l|}{Number of outputs}    & 4                                                                                     \\ \hline
\multicolumn{1}{|l|}{Optimizer}            & Adam                                                                                  \\ \hline
\multicolumn{1}{|l|}{Learning rate}        & 0.001                                                                                 \\ \hline
\multicolumn{1}{|l|}{Epochs}               & 300                                                                                   \\ \hline
\multicolumn{1}{|l|}{Batch size}           & 1024                                                                                   \\ \hline

\end{tabular}
\end{center}
\end{table}


\begin{figure}[ht]
  \centering
  \includegraphics[width=1.0\linewidth]{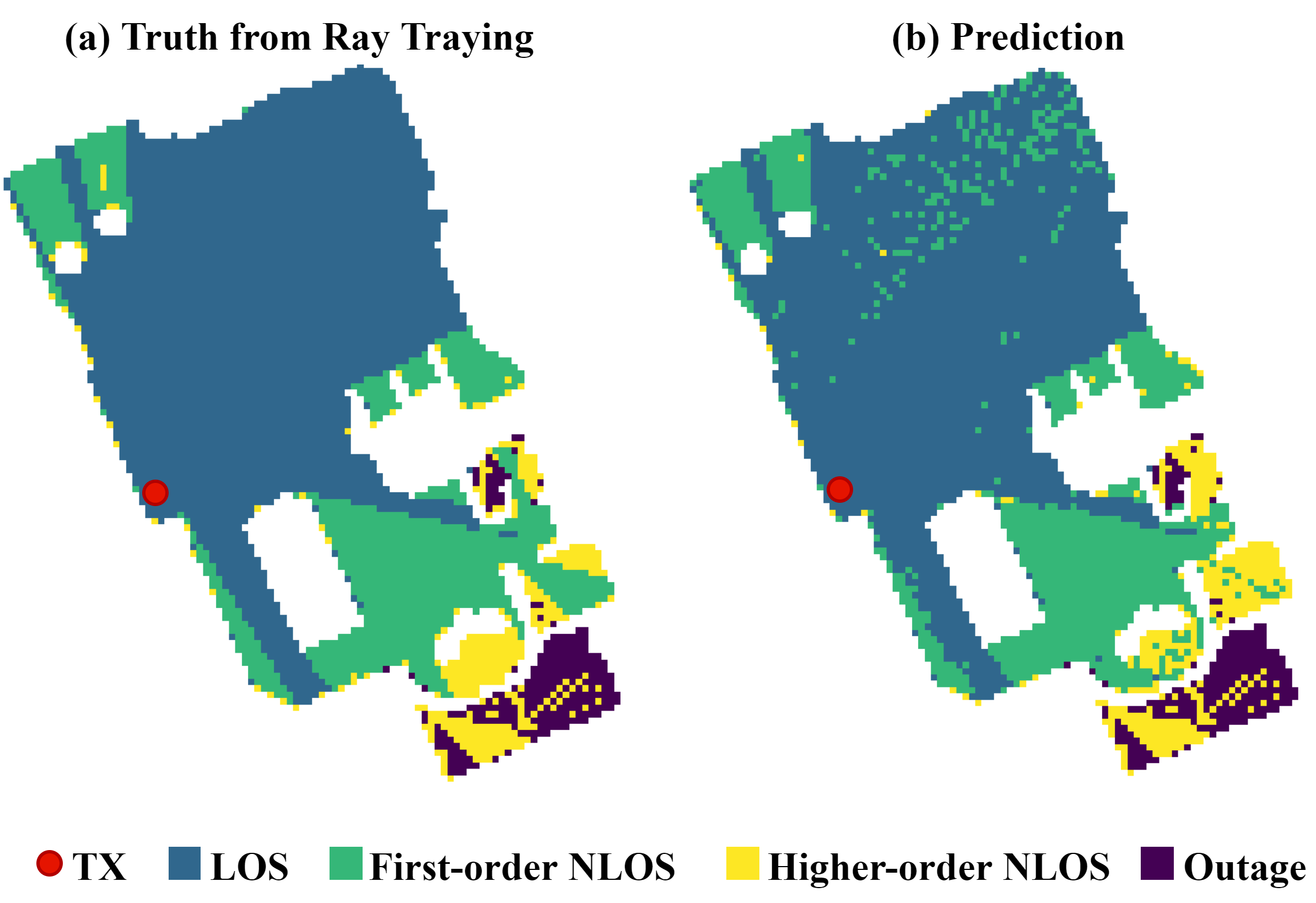}
  \caption{Two Link-States maps. (a) is truth from ray tracing tool and (b) is result of the link-states classification neural network prediction.}
  \label{fig:sl_predict_map}
\end{figure}

To visualize the errors, 
Fig.~\ref{fig:sl_predict_map} plots the
true and predicted link states in a typical environment. 
The true link state (as predicted by ray tracing
and shown on the left)  consistent with common sense.  In particular, 
all RX locations where the TX can be 
directly seen are in the LOS state.
From the right figure, 
we see that the link state is predicted mostly correctly, but there are some errors. 
For the most part, the estimates are conservative
meaning an LOS state is predicted as first-order NLOS, or first-order NLOS predicted as higher-order NLOS.

Finally, it is interesting that one can distinguish first-order
and higher-order LOS so accurately.  Of course, from a 
\emph{single} multi-path component, one cannot know if it 
arrived via one or more reflections.  However, the neural network
demonstrates that 
the \emph{joint} statistics across multiple path components
can reveal the number of reflections with a high accuracy
in completely new environments not part of the training data set.

\subsection{Removing the TX Angle Information}
As described in Section~\ref{sec:sync},
the TX angle estimate 
($\widehat{\Omega}^{\rm tx}_\ell$ in \eqref{eq:pathparam})
may not always be available as it requires 
synchronization and knowledge of the TX codebook
at the RX.  
To study the effect of removing the TX angle,
we re-trained and re-evaluated the 
link classifier performance when the TX
angle is removed.  

\begin{figure}[ht]
  \centering
  \includegraphics[width=1.0\linewidth]{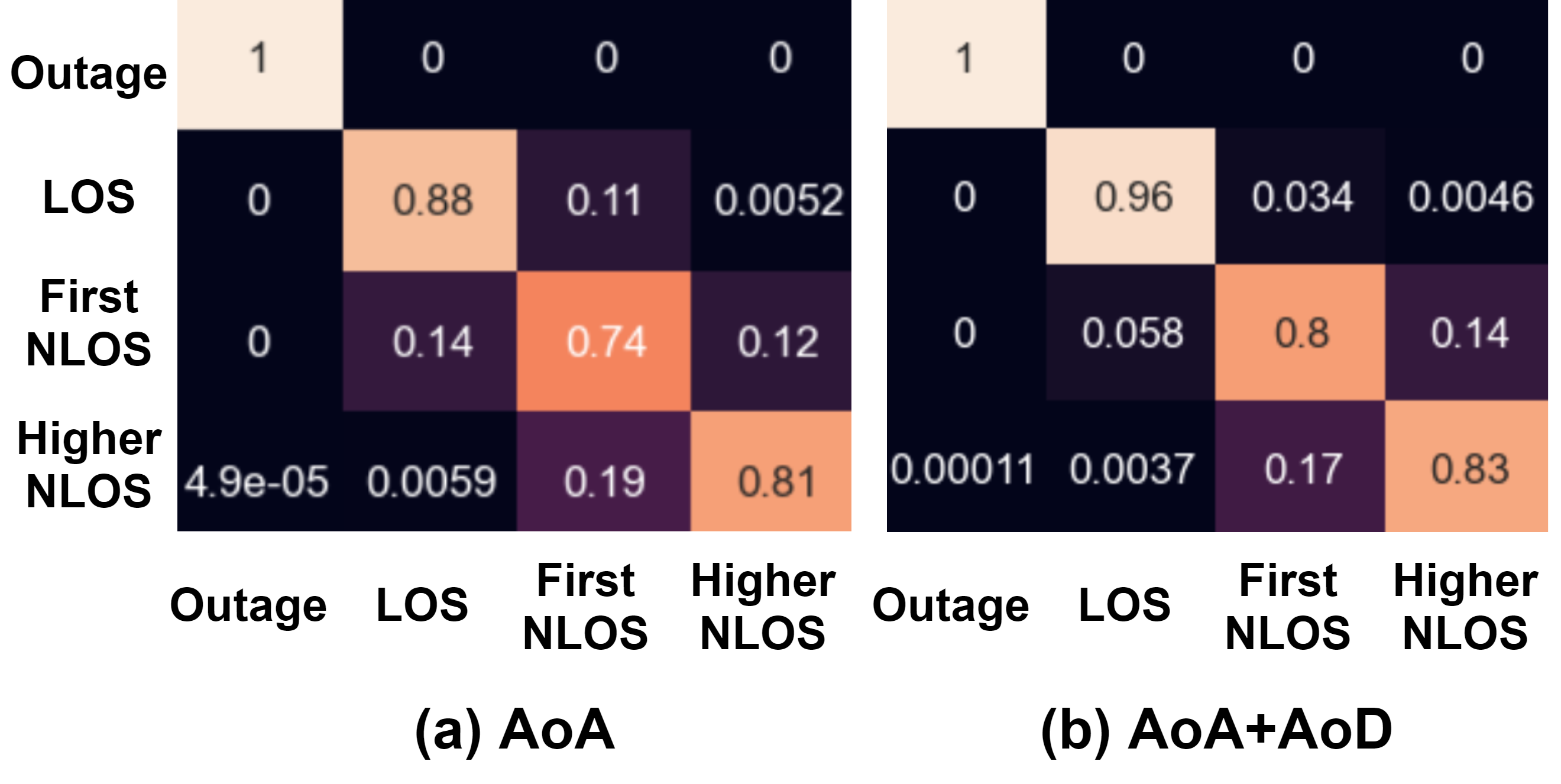}
  \caption{The confusion matrix of
    the validation data set on
    only AoA method and AoA$+$AoD method.
    AoA$+$AoD method improves the classification
    performance in 
    LOS, first-order NLOS, and higer-order NLOS links.}
  \label{fig:confusion_matrix}
\end{figure}

Fig.\ref{fig:confusion_matrix} plots 
the confusion matrix of the validation data set 
using only RX angle (AoA) and RX and TX angle
(AoA$+$AoD).
With the exception of a very small 
difference in the outage state, 
the accuracy of using AoA degrades performance
somewhat.  For example, without using the AoD
estimate, 
the accuracy of LOS state
decreases from $96\%$ to $88\%$.
Furthermore, the AoA-only method incorrectly classifies $25\%$ of the LOS links and the first-order NLOS state.  For the remainder
of the paper, we will assume we have
the TX information, although localization 
without TX information is an area of future work.

%% file: navigation.tex
\section{Integration with Active Neural-SLAM}  \label{sec:navigation}

\begin{figure*}[t]
\footnotesize
\centering
\includegraphics[width=0.85 \linewidth]{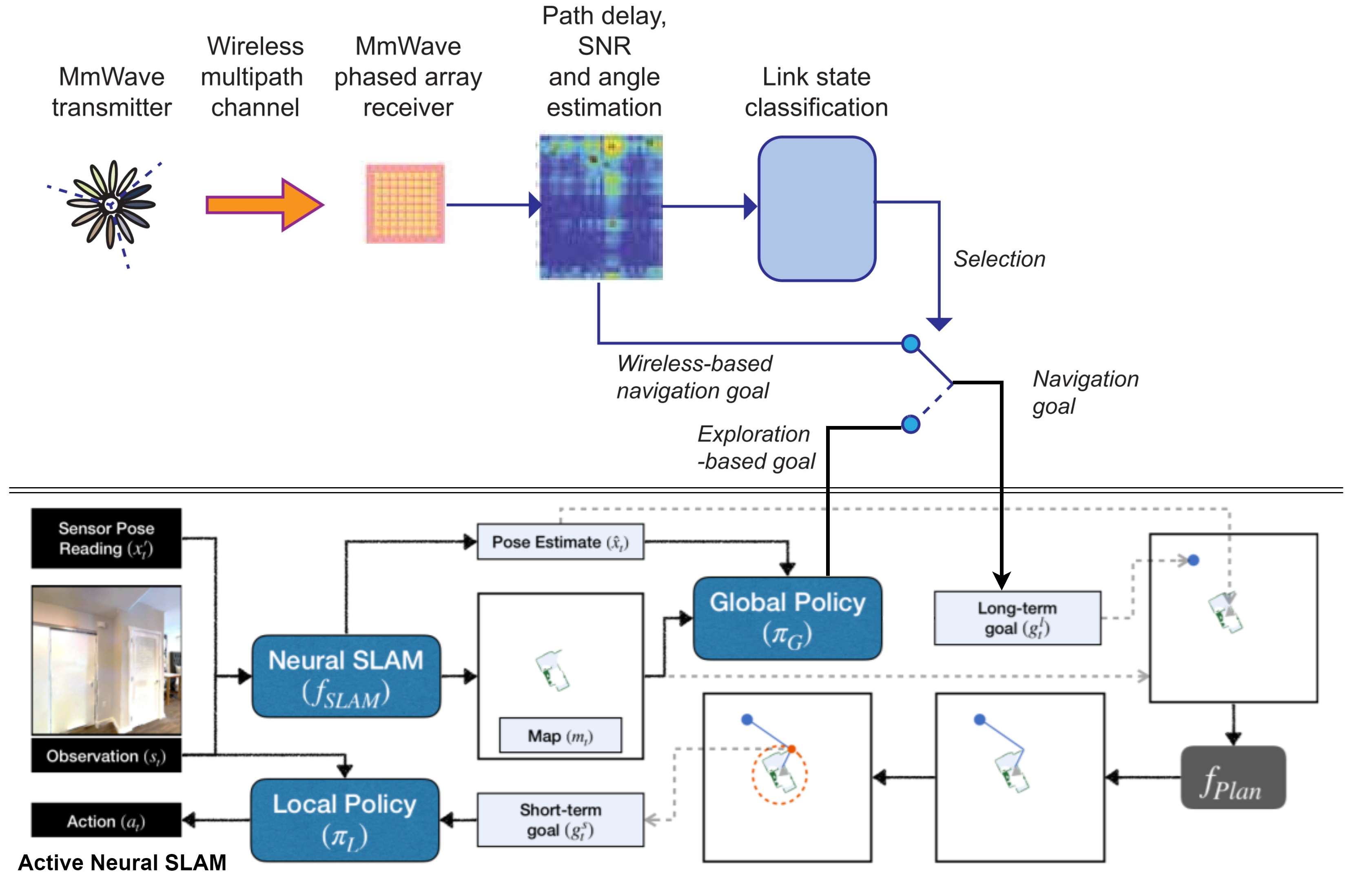}
\caption{MmWave-Based wireless path detection and link
state classification are used to augment the 
Active Neural-SLAM module \cite{chaplot2020learning} by
overwriting the navigation goal from the wireless
path estimation. }
\label{fig:nslam}
\vspace{-2mm}
\end{figure*}

\subsection{Overview of Active Neural-SLAM}
We conclude by integrating the path estimation
and link state classification into the
state-of-the-art Active Neural-SLAM algorithm of \cite{chaplot2020learning} to perform the
target localization.
We first review the basics of the
Active Neural-SLAM algorithm of \cite{chaplot2020learning}.

Neural-SLAM is a powerful new method
for agent navigation in unseen environments,
sometimes referred to as \emph{exploration}. Exploration is a critical task in building intelligent agents. In exploration, the goal of the agent is to explore as much area as it can and as fast as possible. Efficient exploration in a large environment requires the agent to effectively map and memorize the already seen environment, estimating its current state/pose and plan the next actions. Traditional machine learning methods to 
learn to explore have been successful,
but these methods generally require end-to-end reinforcement learning (RL) making the task of exploration difficult in large environments and increase the sample complexity. Instead, Devendra \emph{et.~ al.} \cite{chaplot2020learning} recently proposed Active Neural-SLAM (see the lower half of Fig.~\ref{fig:nslam}) which uses structured spatial representations, hierarchical policies and search based planners. The Active Neural-SLAM consists of three components: Neural-SLAM ($f_{SLAM}$), Global Policy ($\pi_{G}$), and Local Policy ($\pi_{L}$). The Neural-SLAM component is trained using a supervised learning approach to predict the top-down 2D map and estimate the agent’s current pose based on the incoming RGB camera images and sensor pose readings. The Global policy is trained using RL with rewards proportional to the increase in coverage. The Global policy takes as input the predicted top-down 2D map and the estimated agent’s pose from the Neural-SLAM component to output a long term goal on the map. Given the long-term goal, predicted top-down map and agent pose estimate, a Fast Marching planner computes the shortest path from the current agent location to the long-term goal in the unexplored area on the map. A short-term goal is generated on this planned path at a farthest point within \SI{0.25}{m} of the agent’s current location. The local policy is trained using imitation learning, wherein it takes input as RGB camera images and short term goal to output a navigational action. Active Neural-SLAM uses AI Habitat Simulator and is shown to out-perform existing end-to-end RL methods and other baselines on Gibson and Matterport datasets for exploration tasks. 

\subsection{Integration of Wireless Information
for Target Discovery}

The modular structure of the Active Neural-SLAM 
framework provides a simple method
to add wireless information for the target 
discovery problem.  As mentioned in the Introduction, suppose the mobile agent must locate and navigate to target.  The target's location 
is not known, but it broadcasts positioning signals.
To navigate to the target we can augment 
the Active Neural-SLAM module with the wireless
signal processing, shown in 
the top half of Fig.~\ref{fig:nslam}.

As described above, the mobile agent
will have a wireless receiver and can estimate
the wireless paths and their angles of arrivals (AoAs).
If the target was known to be in a LOS state,
and the AoA was reliable, the agent 
could simply follow the AoA of the strongest
path.  In Fig.~\ref{fig:nslam}, this 
navigation goal is referred to as the 
"wireless-based navigation goal".  

Of course, the link state and the path 
estimate are not known \emph{a priori}
by the mobile agent.  We thus propose to 
use the link state classification along with
the estimated SNR of the strongest
path to make a decision on whether to use
the wireless-based navigation goal or not.
If the wireless-based navigation goal is selected,
it can simply overwrite the
navigation goal in the Active Neural-SLAM module.
If, on the other hand, the wireless-based
navigation goal is considered
unreliable, the mobile agent can use
the exploration-based goal from the original global policy.
This selection concept is illustrated
in Fig.~\ref{fig:nslam}.

The resulting algorithm structure thus
attempts to use the wireless based AoA when it 
is accurate, but goes back to the exploration
when the wireless signals are not reliable
or not likely to point into a useful direction
to the target.

We consider three possible selection
algorithms for determining whether or not
to use
the estimated AoA from the wireless detection:
\begin{itemize}
    \item \emph{AoA based on SNR only}: 
    The robot follows the AoA of the highest SNR path if the path SNR is above some threshold
    in any link state.
    Otherwise, the robot follows the goal from 
    Active Neural-SLAM map exploration.
    
    \item \emph{AoA when LOS}: 
    The robot follows the estimated
    AoA when the strongest path is in a LOS state
    and the SNR is above the threshold.
    
    \item \emph{AoA when LOS or First-order NLOS}:
    The robot follows the estimated
    AoA when the strongest path is in a LOS state
    or first-order NLOS 
    and the SNR is above the threshold.
\end{itemize}
In the simulations below we use a minimum 
SNR threshold of \SI{10}{dB} for all three options,
which we found to  give the best results.
Note that the second and third options
require link state classification
and the third option requires differentiation
between first-order and higher-order NLOS.

To better evaluate the performance of our proposed \emph{AoA when LOS or First-order NLOS} wireless assisted algorithm, we use a state-of-the-art completely computer vision-based navigation algorithm which calls the \emph{Visual LOS CV-Based SLAM} navigation algorithm. 
The \emph{Visual LOS CV-Based SLAM} navigation algorithm is defined as that: 
the robot first starts to use the Active Neural-SLAM algorithm [11] to explore the unknown environment, 
and as soon as the target appears in the robot’s field of view, 
we assume that the robot uses a computer vision technology to identify and locate the target with 100\% accuracy, 
and then finally the robot sets the goal on the position of the target and navigates itself.

\subsection{Evaluation}
To evaluate different algorithms, we compared them
to a \emph{baseline} algorithm where
the robot knows the position (coordinates) of the target (TX) and always sets walking-goal at the TX.
The baseline is essentially an oracle that
provides an upper bound on the performance
of any algorithm that does not know the
TX location.

The baseline algorithm also provides a 
simple classification of the environments.
Twenty validation maps with a total of 193 independently validated tests were run. Based on the number of steps taken by the robot to reach the destination of the baseline algorithm, we classified the tests into three difficulty levels: easy, moderate, and hard. And there are 52 easy tests, 76 moderate tests, and 65 hard tests.  We will present the results separately
for these cases.

In order to evaluate the performance of an algorithm, we rely on two criteria: arrival 
success rate and arrival time.
The robot is defined to have succeeded at arriving at the destination if it arrives at the TX location within 1000 steps.
The arrival success rate (sometimes, simply
arrival rate) is the fraction of the number of time of
the robot succeeds in arriving.  
In the case of a successful arrival, the arrival
time is measured relative to the baseline algorithm.
As a result, an algorithm with the lowest arrival time and highest arrival success rate is preferable.

\begin{figure}[t]
  \centering
  \includegraphics[width=1.0\linewidth]{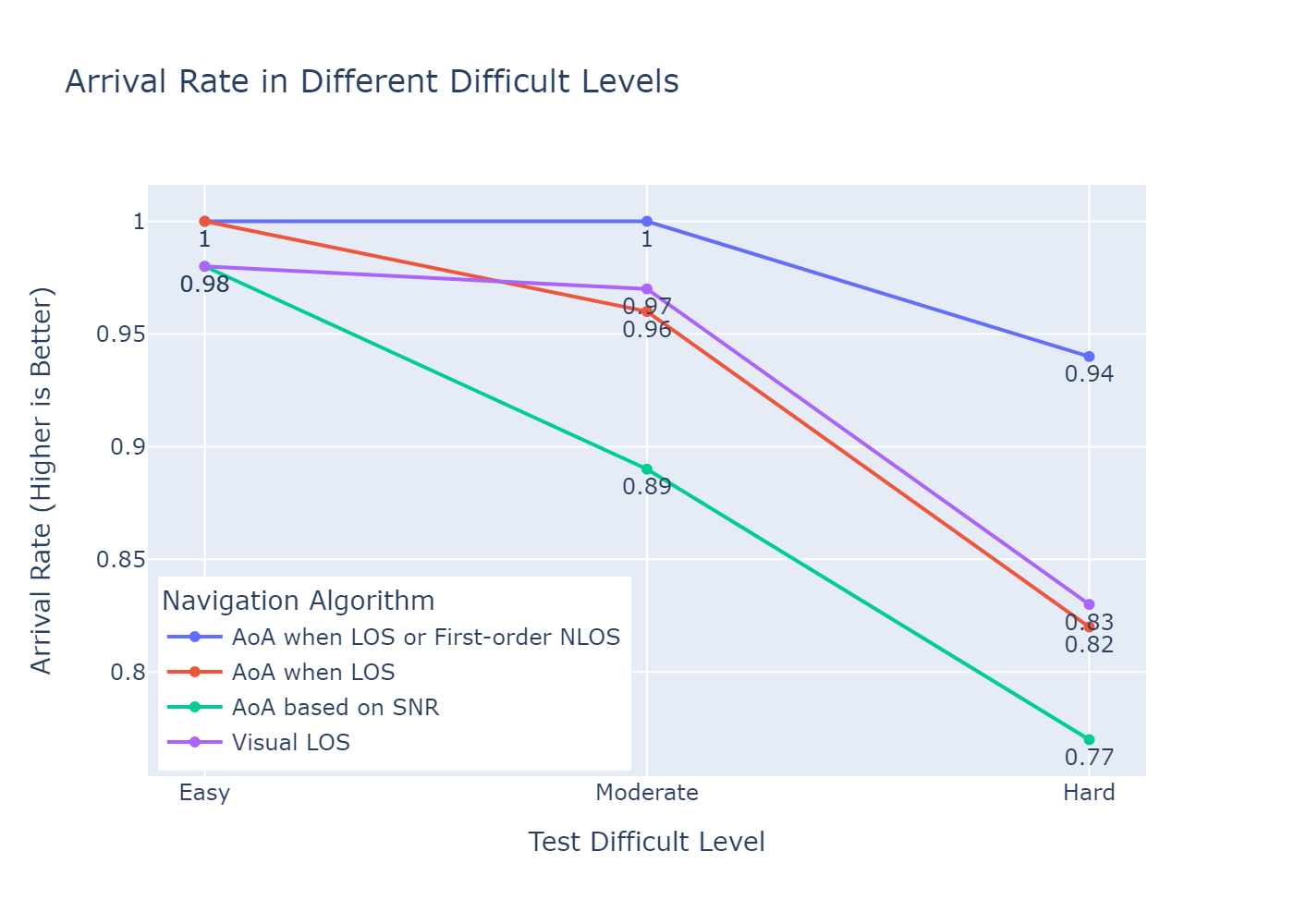}
  \caption{The arrival success rate of three algorithms in easy, moderate, and hard environments.
  Our proposed \emph{AoA when LOS or First-order NLOS} wireless assisted algorithm has the highest arrival success rate in all test levels, and it beats the stat-of-art completely computer vision-based navigation algorithm.}
  \label{fig:arrival_rate}
\end{figure}

\begin{figure}[t]
  \centering
  \includegraphics[width=0.9\linewidth]{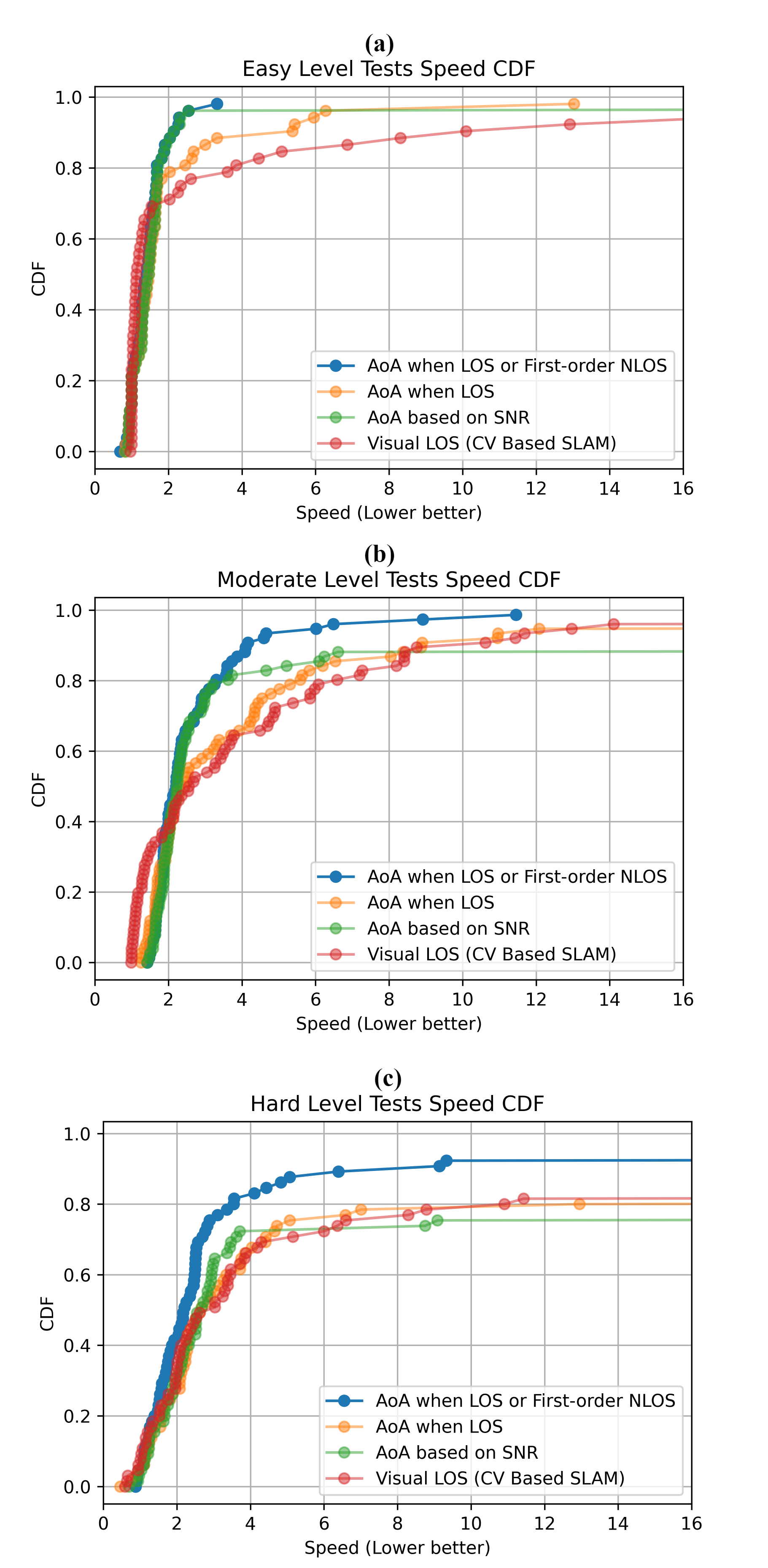}
  \caption{Three cumulative distribution function (CDF) plots show the arrival speed in easy, moderate, and hard difficult level.
  At the moderate and hard difficulty levels, our proposed \emph{AoA when LOS or First-order NLOS} wireless assisted algorithm performs most effectively. 
  The results demonstrate the effectiveness of the four-way link-state classification neural network in improving the robot indoor navigation problem.}
  \label{fig:arrival_speed}
\end{figure}

Fig. \ref{fig:arrival_rate} plots the arrival 
success rates for the four algorithms in the easy, moderate and hard environments.  
We see that without the assistance of the link-state classification neural network, the robot using the \emph{AoA based on SNR only} algorithm gets stuck and is unable to succeed in 11 percent of the moderate tests and 23 percent of the difficult tests.
At the same time, the \emph{AoA when LOS or First-order NLOS} algorithm is superior to \emph{AoA when LOS}.
This result suggests that the four-way
link state classification provides a more
robust metric for using the wireless information.
The reason for this is that even though the estimation error of AoA in the first-order NLOS is larger than the LOS, the robot can still use the AoA information to navigate itself and avoid getting stuck with high probability.
As we use the beam sweeping double directional path estimation for both AoA and AoD, our link-states classification neural network shows high accuracy in identifying LOS, first-order NLOS, and higher-order NLOS.
Since the estimation error of AoA varies greatly under the three different link states, as shown in Fig. \ref{fig:aoa_est_err}, we consider that using the angular information in the wireless signal to help navigation in the case of LOS and first-order NLOS is the optimal solution which can reach a highest arrival rate.
In addition, the \emph{AoA when LOS or First-order NLOS} wireless assisted algorithm is the only one that defeats the state-of-the-art completely computer vision-based navigation algorithm in all three difficult levels.
As a result, the robot can utilize the information provided by the wireless signal to the maximum extent without being affected by the noise in the wireless communications that may cause the robot to move in the wrong direction.

In order to analyze the arrival time, Fig.~\ref{fig:arrival_speed} plots the cumulative distribution functions (CDFs) of the
arrival time for the different algorithms in 
the three difficulty levels. Instances where
the algorithm did not succeed as arriving
are treated as a big arrival time.
Note that the arrival times
are relative to the baseline with an arrival
time of one being optimal.

We see that in all three difficult levels,
the arrival time of \emph{AoA based on SNR only} and \emph{AoA when LOS only} algorithms are slightly better than the \emph{Visual LOS CV-Based SLAM} navigation algorithm,
which means that without a link-state classification or with a three-way (LOS, NLOS, outage) link-state classifier, the mmWave wireless system have very limited optimization of the state-of-the-art completely computer vision-based navigation algorithm.
However, with our new four-way link-state classifier (LOS, fisrt-order NLOS, higher-order NLOS, outage), the \emph{AoA when LOS or First-order NLOS} wireless assisted algorithm performs most effectively in moderate and hard level tests.

\begin{figure}[t]
  \centering
  \includegraphics[width=1.0\linewidth]{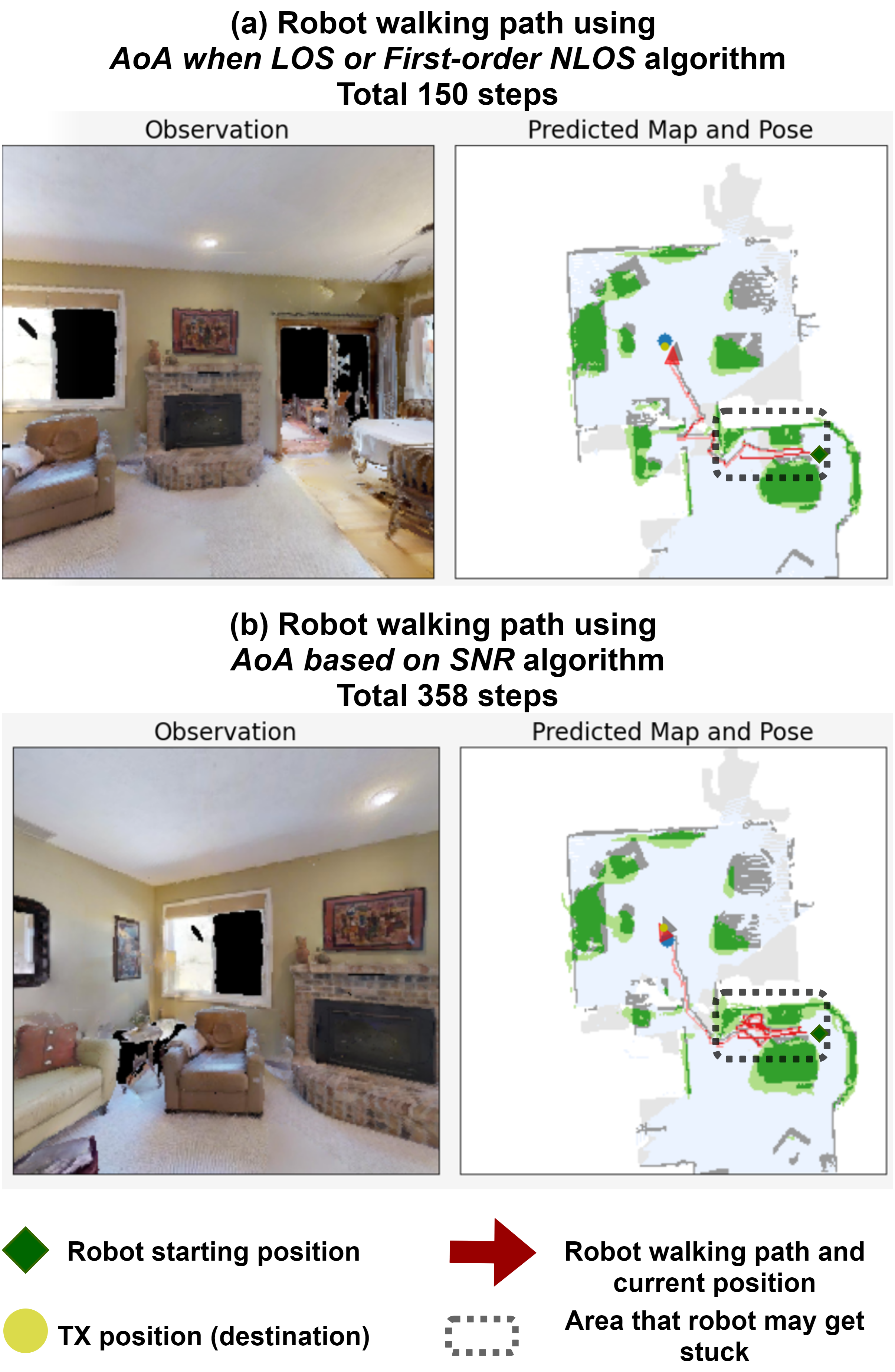}
  \caption{An example of two different robot walling paths are generated in a test case. In (a),
  the robot uses the \emph{AoA when LOS or First-order NLOS} and spends 150 steps to arrive the TX. 
  In (b), the robot uses the \emph{AoA based on SNR} and spends 358  steps to reach the TX.
  The area framed by the black dashed line shows the difference between the two algorithms.}
  \label{fig:example_lsaoa_better}
\end{figure}

\subsection{Representative Example}
It is useful to look at a representative example,
to visualize the problems with navigation 
without the link state classification.  
Fig. \ref{fig:example_lsaoa_better}
shows the predicted map and route followed by
robot with  \emph{AoA when LOS or First-order NLOS} on the top panel 
and \emph{AoA based on SNR} on the bottom
panel.  The robot is started at the same location
in both cases.  In this case, the  starting point are in higher-order NLOS link-state.
In the AoA based on SNR algorithm, the robot
follows the AoA.  Since it is a high-order LOS,
the direction is not valuable even if it is
reliably detected.  As a result, the robot spends a significant amount of time in the area framed by the black dashed line in Fig. \ref{fig:example_lsaoa_better}(b).
In contrast, the \emph{AoA when LOS or First-order NLOS} algorithm successfully detects
that the location is a high-order NLOS state.
The robot then ignores the AoA initially and
instead relies on the Active Neural-SLAM exploration
which rapidly takes it into a new area.  

%% file: conclusion.tex
\section{Conclusions}
\label{sec:conclusion}

We have considered using mmWave-based positioning
for target localization in unknown environment.  
This problem is relatively new
in the context of mmWave-based localization.
We have considered an algorithm that uses
advanced wireless path estimation and link state
classification methods and integrated these
into a state-of-the-art 
neural SLAM module for complete robotic
navigation.  Our results demonstrate that
target navigation is possible, 
even in challenging new environments.
In particular, the results demonstrate the importance
of accurately determining the link state
including differentiating between first-order
and higher-order NLOS states.

The proposed method is, however, a starting  point
and there are several open avenues for future work.
Most obviously, the wireless and visual sensor
data are not currently processed jointly: the proposed
algorithm uses one or the other at any given time. 
Moreover, in the proposed approach,
the wireless information is not used for SLAM,
only for localization.  There is now a large body
of work on mmWave SLAM \cite{guidi2014millimeter,palacios2018communication}
and one avenue of future work is to 
incorporate these or other methods.
Indeed, the camera and RF signals represent a
``digital twin" from one another with significant
possibilities for joint processing.    
More broadly, there is a rich interplay between visual
processing, wireless sensing and navigation.
To assist future work in this area, we have developed a fully open source data set \cite{mmwRobotNav}
that integrates the robotic simulation,
camera data and wireless coverage.  

%% file: main.bbl
\begin{thebibliography}{10}
\providecommand{\url}[1]{#1}
\csname url@samestyle\endcsname
\providecommand{\newblock}{\relax}
\providecommand{\bibinfo}[2]{#2}
\providecommand{\BIBentrySTDinterwordspacing}{\spaceskip=0pt\relax}
\providecommand{\BIBentryALTinterwordstretchfactor}{4}
\providecommand{\BIBentryALTinterwordspacing}{\spaceskip=\fontdimen2\font plus
\BIBentryALTinterwordstretchfactor\fontdimen3\font minus
  \fontdimen4\font\relax}
\providecommand{\BIBforeignlanguage}[2]{{%
\expandafter\ifx\csname l@#1\endcsname\relax
\typeout{** WARNING: IEEEtran.bst: No hyphenation pattern has been}%
\typeout{** loaded for the language `#1'. Using the pattern for}%
\typeout{** the default language instead.}%
\else
\language=\csname l@#1\endcsname
\fi
#2}}
\providecommand{\BIBdecl}{\relax}
\BIBdecl

\bibitem{guerra2015position}
A.~Guerra, F.~Guidi, and D.~Dardari, ``Position and orientation error bound for
  wideband massive antenna arrays,'' in \emph{2015 IEEE International
  Conference on Communication Workshop (ICCW)}.\hskip 1em plus 0.5em minus
  0.4em\relax IEEE, 2015, pp. 853--858.

\bibitem{shahmansoori20155g}
A.~Shahmansoori, G.~E. Garcia, G.~Destino, G.~Seco-Granados, and H.~Wymeersch,
  ``5g position and orientation estimation through millimeter wave mimo,'' in
  \emph{2015 IEEE Globecom Workshops (GC Wkshps)}.\hskip 1em plus 0.5em minus
  0.4em\relax IEEE, 2015, pp. 1--6.

\bibitem{guidi2014millimeter}
F.~Guidi, A.~Guerra, and D.~Dardari, ``Millimeter-wave massive arrays for
  indoor slam,'' in \emph{2014 IEEE International Conference on Communications
  Workshops (ICC)}.\hskip 1em plus 0.5em minus 0.4em\relax IEEE, 2014, pp.
  114--120.

\bibitem{3GPP38885}
{3GPP Technical Report 38.885}, ``Study on {NR} positioning support ({R}elease
  16),'' Mar. 2019.

\bibitem{3GPP38857}
{3GPP Technical Report 38.857}, ``Study on {NR} positioning enhancements
  ({R}elease 17),'' Mar. 2021.

\bibitem{3GPP21916}
{3GPP Technical Report 21.916}, ``Release description; {R}elease 16 ({R}elease
  16),'' Jun. 2021.

\bibitem{keating2019overview}
R.~Keating, M.~S{\"a}ily, J.~Hulkkonen, and J.~Karjalainen, ``{Overview of
  positioning in 5G new radio},'' in \emph{2019 16th International Symposium on
  Wireless Communication Systems (ISWCS)}.\hskip 1em plus 0.5em minus
  0.4em\relax IEEE, 2019, pp. 320--324.

\bibitem{dwivedi2021positioning}
S.~Dwivedi, R.~Shreevastav, F.~Munier, J.~Nygren, I.~Siomina, Y.~Lyazidi,
  D.~Shrestha, G.~Lindmark, P.~Ernstr{\"o}m, E.~Stare \emph{et~al.},
  ``{Positioning in 5G networks},'' \emph{arXiv preprint arXiv:2102.03361},
  2021.

\bibitem{zwirello2012uwb}
L.~Zwirello, T.~Schipper, M.~Harter, and T.~Zwick, ``Uwb localization system
  for indoor applications: concept, realization and analysis,'' \emph{Journal
  of Electrical and Computer Engineering}, vol. 2012, 2012.

\bibitem{zhang2008high}
Y.~Zhang, A.~K. Brown, W.~Q. Malik, and D.~J. Edwards, ``High resolution 3-d
  angle of arrival determination for indoor uwb multipath propagation,''
  \emph{IEEE Transactions on Wireless Communications}, vol.~7, no.~8, pp.
  3047--3055, 2008.

\bibitem{chaplot2020learning}
D.~S. Chaplot, D.~Gandhi, S.~Gupta, A.~Gupta, and R.~Salakhutdinov, ``{Learning
  to explore using active neural SLAM},'' \emph{arXiv preprint
  arXiv:2004.05155}, 2020.

\bibitem{slezak2018empirical}
C.~Slezak, V.~Semkin, S.~Andreev, Y.~Koucheryavy, and S.~Rangan, ``{Empirical
  effects of dynamic human-body blockage in 60 GHz communications},''
  \emph{IEEE Communications Magazine}, vol.~56, no.~12, pp. 60--66, 2018.

\bibitem{Rappaport2014-mmwbook}
T.~S. Rappaport, R.~W. {Heath Jr.}, R.~C. Daniels, and J.~N. Murdock,
  \emph{Millimeter Wave Wireless Communications}.\hskip 1em plus 0.5em minus
  0.4em\relax Pearson Education, 2014.

\bibitem{guvenc2007nlos}
I.~Guvenc, C.-C. Chong, and F.~Watanabe, ``Nlos identification and mitigation
  for uwb localization systems,'' in \emph{2007 IEEE Wireless Communications
  and Networking Conference}.\hskip 1em plus 0.5em minus 0.4em\relax IEEE,
  2007, pp. 1571--1576.

\bibitem{venkatesh2007non}
S.~Venkatesh and R.~Buehrer, ``Non-line-of-sight identification in
  ultra-wideband systems based on received signal statistics,'' \emph{IET
  Microwaves, Antennas \& Propagation}, vol.~1, no.~6, pp. 1120--1130, 2007.

\bibitem{khodjaev2010survey}
J.~Khodjaev, Y.~Park, and A.~S. Malik, ``Survey of nlos identification and
  error mitigation problems in uwb-based positioning algorithms for dense
  environments,'' \emph{annals of telecommunications-annales des
  t{\'e}l{\'e}communications}, vol.~65, no.~5, pp. 301--311, 2010.

\bibitem{wang2018pursuance}
D.~Wang, M.~Fattouche, and X.~Zhan, ``Pursuance of mm-level accuracy: Ranging
  and positioning in mmwave systems,'' \emph{IEEE Systems Journal}, vol.~13,
  no.~2, pp. 1169--1180, 2018.

\bibitem{lu2020positioning}
Y.~Lu, M.~Koivisto, J.~Talvitie, M.~Valkama, and E.~S. Lohan,
  ``{Positioning-aided 3D beamforming for enhanced communications in mmWave
  mobile networks},'' \emph{IEEE Access}, vol.~8, pp. 55\,513--55\,525, 2020.

\bibitem{kim20185g}
H.~Kim, H.~Wymeersch, N.~Garcia, G.~Seco-Granados, and S.~Kim, ``{5G mmWave
  vehicular tracking},'' in \emph{2018 52nd Asilomar Conference on Signals,
  Systems, and Computers}.\hskip 1em plus 0.5em minus 0.4em\relax IEEE, 2018,
  pp. 541--547.

\bibitem{zhou2017low}
Z.~Zhou, J.~Fang, L.~Yang, H.~Li, Z.~Chen, and R.~S. Blum, ``{Low-rank tensor
  decomposition-aided channel estimation for millimeter wave MIMO-OFDM
  systems},'' \emph{IEEE Journal on Selected Areas in Communications}, vol.~35,
  no.~7, pp. 1524--1538, 2017.

\bibitem{wen2018tensor}
F.~Wen, N.~Garcia, J.~Kulmer, K.~Witrisal, and H.~Wymeersch, ``{Tensor
  decomposition based beamspace ESPRIT for millimeter wave MIMO channel
  estimation},'' in \emph{2018 IEEE Global Communications Conference
  (GLOBECOM)}.\hskip 1em plus 0.5em minus 0.4em\relax IEEE, 2018, pp. 1--7.

\bibitem{raghavan2019statistical}
V.~Raghavan, L.~Akhoondzadeh-Asl, V.~Podshivalov, J.~Hulten, M.~A. Tassoudji,
  O.~H. Koymen, A.~Sampath, and J.~Li, ``Statistical blockage modeling and
  robustness of beamforming in millimeter-wave systems,'' \emph{IEEE
  Transactions on Microwave Theory and Techniques}, vol.~67, no.~7, pp.
  3010--3024, 2019.

\bibitem{xia2020multi}
W.~Xia, V.~Semkin, M.~Mezzavilla, G.~Loianno, and S.~Rangan, ``Multi-array
  designs for mmwave and sub-thz communication to uavs,'' in \emph{2020 IEEE
  21st International Workshop on Signal Processing Advances in Wireless
  Communications (SPAWC)}.\hskip 1em plus 0.5em minus 0.4em\relax IEEE, 2020,
  pp. 1--5.

\bibitem{guerra2018single}
A.~Guerra, F.~Guidi, and D.~Dardari, ``Single-anchor localization and
  orientation performance limits using massive arrays: Mimo vs. beamforming,''
  \emph{IEEE Transactions on Wireless Communications}, vol.~17, no.~8, pp.
  5241--5255, 2018.

\bibitem{huang2020machine}
C.~Huang, A.~F. Molisch, R.~He, R.~Wang, P.~Tang, B.~Ai, and Z.~Zhong,
  ``Machine learning-enabled los/nlos identification for mimo systems in
  dynamic environments,'' \emph{IEEE Transactions on Wireless Communications},
  vol.~19, no.~6, pp. 3643--3657, 2020.

\bibitem{heath2016overview}
R.~W. Heath, N.~Gonzalez-Prelcic, S.~Rangan, W.~Roh, and A.~M. Sayeed, ``{An
  overview of signal processing techniques for millimeter wave MIMO systems},''
  \emph{IEEE journal of selected topics in signal processing}, vol.~10, no.~3,
  pp. 436--453, 2016.

\bibitem{xia2018gibson}
F.~Xia, A.~R. Zamir, Z.~He, A.~Sax, J.~Malik, and S.~Savarese, ``Gibson env:
  Real-world perception for embodied agents,'' in \emph{Proceedings of the IEEE
  Conference on Computer Vision and Pattern Recognition}, 2018, pp. 9068--9079.

\bibitem{savva2019habitat}
M.~Savva, A.~Kadian, O.~Maksymets, Y.~Zhao, E.~Wijmans, B.~Jain, J.~Straub,
  J.~Liu, V.~Koltun, J.~Malik \emph{et~al.}, ``{Habitat: A platform for
  embodied AI research},'' in \emph{Proceedings of the IEEE/CVF International
  Conference on Computer Vision}, 2019, pp. 9339--9347.

\bibitem{Remcom}
``{Remcom},'' available on-line at {\tt https://www.remcom.com/}.

\bibitem{mmwRobotNav}
M.~Yin, ``{mmwRobotNav} git hub repository,'' available on-line at {\tt
  https://github.com/nyu-wireless/mmwRobotNav}.

\bibitem{heath2018foundations}
R.~W. Heath~Jr. and A.~Lozano, \emph{Foundations of {MIMO}
  Communication}.\hskip 1em plus 0.5em minus 0.4em\relax Cambridge University
  Press, 2018.

\bibitem{kelley1993array}
D.~F. Kelley and W.~L. Stutzman, ``Array antenna pattern modeling methods that
  include mutual coupling effects,'' \emph{IEEE Transactions on antennas and
  propagation}, vol.~41, no.~12, pp. 1625--1632, 1993.

\bibitem{song2015codebook}
J.~Song, J.~Choi, and D.~J. Love, ``Codebook design for hybrid beamforming in
  millimeter wave systems,'' in \emph{2015 IEEE International Conference on
  Communications (ICC)}.\hskip 1em plus 0.5em minus 0.4em\relax IEEE, 2015, pp.
  1298--1303.

\bibitem{3GPP38901}
{3GPP Technical Report 38.901}, ``Study on channel model for frequencies from
  0.5 to 100 {GHz} ({R}elease 16),'' Dec. 2019.

\bibitem{grasedyck2013literature}
L.~Grasedyck, D.~Kressner, and C.~Tobler, ``A literature survey of low-rank
  tensor approximation techniques,'' \emph{GAMM-Mitteilungen}, vol.~36, no.~1,
  pp. 53--78, 2013.

\bibitem{giordani2018tutorial}
M.~Giordani, M.~Polese, A.~Roy, D.~Castor, and M.~Zorzi, ``{A Tutorial on Beam
  Management for 3GPP NR at mmWave Frequencies},'' \emph{IEEE Communications
  Surveys \& Tutorials}, vol.~21, no.~1, pp. 173--196, 2018.

\bibitem{yu2020meta}
T.~Yu, D.~Quillen, Z.~He, R.~Julian, K.~Hausman, C.~Finn, and S.~Levine,
  ``Meta-world: A benchmark and evaluation for multi-task and meta
  reinforcement learning,'' in \emph{Conference on Robot Learning}.\hskip 1em
  plus 0.5em minus 0.4em\relax PMLR, 2020, pp. 1094--1100.

\bibitem{xia2020interactive}
F.~Xia, W.~B. Shen, C.~Li, P.~Kasimbeg, M.~E. Tchapmi, A.~Toshev,
  R.~Mart{\'\i}n-Mart{\'\i}n, and S.~Savarese, ``Interactive gibson benchmark:
  A benchmark for interactive navigation in cluttered environments,''
  \emph{IEEE Robotics and Automation Letters}, vol.~5, no.~2, pp. 713--720,
  2020.

\bibitem{taha2021millimeter}
A.~Taha, Q.~Qu, S.~Alex, P.~Wang, W.~L. Abbott, and A.~Alkhateeb, ``Millimeter
  wave mimo-based depth maps for wireless virtual and augmented reality,''
  \emph{IEEE Access}, vol.~9, pp. 48\,341--48\,363, 2021.

\bibitem{khawaja2017uav}
W.~Khawaja, O.~Ozdemir, and I.~Guvenc, ``{UAV Air-to-Ground Channel
  Characterization for mmWave Systems},'' in \emph{Proc.\ IEEE VTC-Fall}, 2017.

\bibitem{xia2020generative}
W.~Xia, S.~Rangan, M.~Mezzavillla, A.~Lozano, G.~Geraci, V.~Semkin, and
  G.~Loianno, ``Generative neural network channel modeling for millimeter-wave
  uav communication,'' \emph{arXiv preprint arXiv:2012.09133}, 2020.

\bibitem{habitat19iccv}
M.~Savva, A.~Kadian, O.~Maksymets, Y.~Zhao, E.~Wijmans, B.~Jain, J.~Straub,
  J.~Liu, V.~Koltun, J.~Malik, D.~Parikh, and D.~Batra, ``Habitat: {A}
  {P}latform for {E}mbodied {AI} {R}esearch,'' in \emph{Proceedings of the
  IEEE/CVF International Conference on Computer Vision (ICCV)}, 2019.

\bibitem{szot2021habitat}
A.~Szot, A.~Clegg, E.~Undersander, E.~Wijmans, Y.~Zhao, J.~Turner, N.~Maestre,
  M.~Mukadam, D.~Chaplot, O.~Maksymets, A.~Gokaslan, V.~Vondrus, S.~Dharur,
  F.~Meier, W.~Galuba, A.~Chang, Z.~Kira, V.~Koltun, J.~Malik, M.~Savva, and
  D.~Batra, ``Habitat 2.0: Training home assistants to rearrange their
  habitat,'' \emph{arXiv preprint arXiv:2106.14405}, 2021.

\bibitem{dahlman20205g}
E.~Dahlman, S.~Parkvall, and J.~Skold, \emph{{5G NR: The next generation
  wireless access technology}}.\hskip 1em plus 0.5em minus 0.4em\relax Academic
  Press, 2020.

\bibitem{gu2021development}
X.~Gu, A.~Paidimarri, B.~Sadhu, C.~Baks, S.~Lukashov, M.~Yeck, Y.~Kwark,
  T.~Chen, G.~Zussman, I.~Seskar \emph{et~al.}, ``{Development of a compact
  28-GHz software-defined phased array for a city-scale wireless research
  testbed},'' in \emph{Proc. IEEE Int. Microw. Symp.(IMS)}, 2021, pp.
  1885--1889.

\bibitem{8606243}
V.~Raghavan, M.-L. Chi, M.~A. Tassoudji, O.~H. Koymen, and J.~Li, ``Antenna
  placement and performance tradeoffs with hand blockage in millimeter wave
  systems,'' \emph{IEEE Transactions on Communications}, vol.~67, no.~4, pp.
  3082--3096, 2019.

\bibitem{7956180}
R.~Garg and A.~S. Natarajan, ``A 28-ghz low-power phased-array receiver
  front-end with 360° rtps phase shift range,'' \emph{IEEE Transactions on
  Microwave Theory and Techniques}, vol.~65, no.~11, pp. 4703--4714, 2017.

\bibitem{nair2010rectified}
V.~Nair and G.~E. Hinton, ``Rectified linear units improve restricted boltzmann
  machines,'' in \emph{Icml}, 2010.

\bibitem{kingma2014adam}
D.~P. Kingma and J.~Ba, ``Adam: A method for stochastic optimization,''
  \emph{arXiv preprint arXiv:1412.6980}, 2014.

\bibitem{palacios2018communication}
J.~Palacios, G.~Bielsa, P.~Casari, and J.~Widmer, ``Communication-driven
  localization and mapping for millimeter wave networks,'' in \emph{IEEE
  INFOCOM 2018-IEEE Conference on Computer Communications}.\hskip 1em plus
  0.5em minus 0.4em\relax IEEE, 2018, pp. 2402--2410.

\end{thebibliography}
